\documentclass{article}

\PassOptionsToPackage{numbers,compress}{natbib}
\usepackage[preprint]{neurips_2024}

\usepackage[utf8]{inputenc} 
\usepackage[T1]{fontenc}    
\usepackage{hyperref}       
\usepackage{url}            
\usepackage{booktabs}       
\usepackage{amsfonts}       
\usepackage{nicefrac}       
\usepackage{microtype}      
\usepackage{xcolor}         
\usepackage{graphicx,wrapfig}
\usepackage{mathtools}
\usepackage{enumitem}
\usepackage{multirow}

\usepackage{amsmath,amsfonts,bm}
\usepackage{cleveref}

\def\eqref#1{equation~\ref{#1}}

\def\1{\bm{1}}

\DeclareMathOperator*{\E}{\mathbb{E}}
\def\parameters{\{A,\theta\}}

\def\vx{{\bm{x}}}
\def\vy{{\bm{y}}}

\DeclareMathAlphabet{\mathsfit}{\encodingdefault}{\sfdefault}{m}{sl}
\SetMathAlphabet{\mathsfit}{bold}{\encodingdefault}{\sfdefault}{bx}{n}

\def\gA{{\mathcal{A}}}

\def\gD{{\mathcal{D}}}

\def\gM{{\mathcal{M}}}
\def\gN{{\mathcal{N}}}

\def\gS{{\mathcal{S}}}
\def\gT{{\mathcal{T}}}

\def\sI{{\mathbb{I}}}

\newcommand{\R}{\mathbb{R}}

\makeatletter
\newcommand{\newreptheorem}[2]{%
\newenvironment{rep#1}[1]{%
 \def\rep@title{#2 \ref{##1}}%
 \begin{rep@theorem}}%
 {\end{rep@theorem}}}
\makeatother

\newreptheorem{theorem}{Theorem}

\title{Amortized Active Causal Induction with Deep Reinforcement Learning}

\author{%
  \textbf{Yashas Annadani}$^{1,2}$
\quad
\textbf{Panagiotis Tigas}$^{3}$
\quad
\textbf{Stefan Bauer}$^{1,2}$
\quad
\textbf{Adam Foster}
\\
$^1$ Helmholtz AI, Munich \quad $^2$ Technical University of Munich\\ $^3$ OATML, University of Oxford\\
}

\begin{document}

\maketitle

\begin{abstract}
  \looseness=-1 
  We present Causal Amortized Active Structure Learning (CAASL), an active intervention design policy that can select interventions that are adaptive, real-time and that does not require access to the likelihood. 
  This policy, an amortized network based on the transformer, is trained with reinforcement learning on a simulator of the design environment, and a reward function that measures how close the true causal graph is to a causal graph posterior inferred from the gathered data. 
  On synthetic data and a single-cell gene expression simulator, we demonstrate empirically that the data acquired through our policy results in a better estimate of the underlying causal graph than alternative strategies.  Our design policy successfully achieves amortized intervention design on the distribution of the training environment while also generalizing well to distribution shifts in test-time design environments. Further, our policy also demonstrates excellent zero-shot generalization to design environments with dimensionality higher than that during training, and to intervention types that it has not been trained on. 
\end{abstract}

\section{Introduction}

\begin{wrapfigure}{r}{5.0cm}
\vspace{-1cm}
\includegraphics[width=5.0cm]{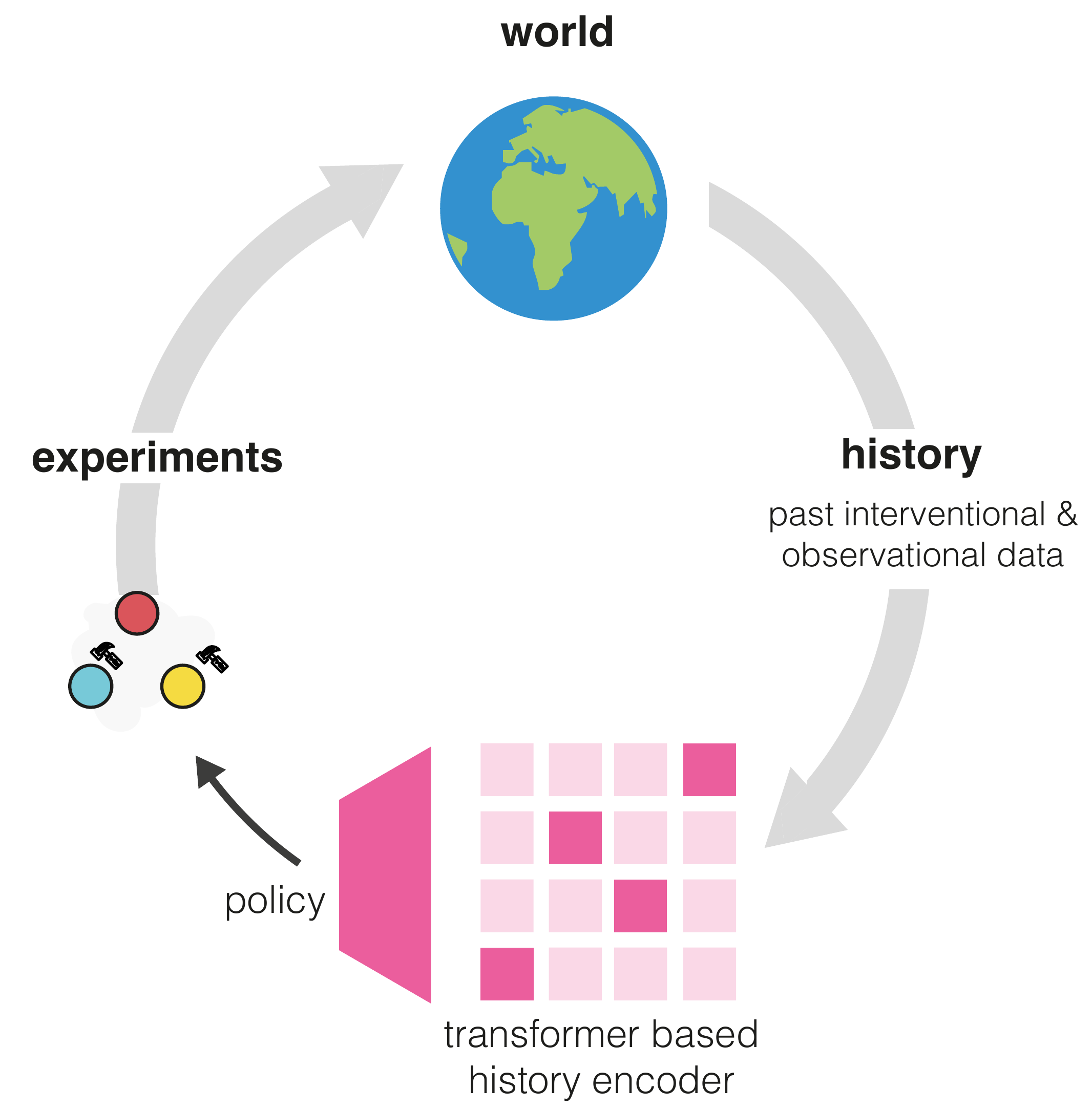}
\caption{Causal Amortized Structure Learning (CAASL) is an active intervention design method that directly proposes the next intervention to perform by just a forward-pass of the transformer based policy.}\label{fig:banner}
\end{wrapfigure} 
Infer, design and experiment is a three step loop in the empirical scientific discovery paradigm. Causal induction (a.k.a.~causal structure learning), the problem of finding causal relationships present in data, also falls under this paradigm when experiments in the form of interventions are permissible~\citep{spirtes2001causation,heinze2018causal}. Causal structure learning has gained increasing importance in empirical sciences, for example in single-cell biology, where perturbation experiments like gene knockouts can be carried out with high-precision~\citep{tejada2023causal}. Such interventions are not only more informative to infer the underlying causal graph than just observational data, but in certain cases essential to go beyond the Markov equivalence class~\citep{peters2017elements}, making the problem of design of interventions both relevant and important. For the problem of structure learning with interventions, however, inference and design both involve significant challenges. For instance, inference of the causal graph from data usually involves search over the space of graphs with a likelihood (usually weighted by a prior) or score function~\citep{annadani2023bayesdag, brouillard2020differentiable,hauser2012characterization},  which is slow and not robust to violations of data generation assumptions~\citep{montagna2024assumption}. The design of informative interventions, on the other hand, utilizes the inferred causal graph 
from existing data to select promising designs and rank them according to a scoring criterion. This scoring criterion is usually based on an approximation of mutual information between the unknown causal graph and the interventional data~\citep{tigas2023differentiable,tigas2022interventions}, which also involves the (interventional) data likelihood. In problems related to empirical sciences where causal discovery is essential, like inferring a gene regulatory network with gene knockouts or knockdowns, the likelihood of the data is typically intractable. While progress has been made in terms of likelihood-free inference of causal graphs~\citep{lorch2022amortized,ke2022learning}, existing intervention design algorithms have been largely restricted to likelihood-based strategies.

With a focus on addressing practical intervention design challenges that arise in empirical sciences like inferring the gene regulatory network, in this work, we propose an intervention design method called CAASL that significantly differs from existing approaches. Instead of following the infer, design and experiment loop, we amortize the intervention design procedure by training a single design network policy, based on the transformer~\citep{vaswani2017attention}, which encodes key design space symmetries. During test-time, our trained policy directly predicts the next intervention to perform by just a forward-pass of the data collected so far, without the need to undergo slow and expensive inference of the causal graph corresponding to that data. We train the transformer policy with Soft Actor-Critic (SAC)~\citep{haarnoja2018soft} to maximize cumulative rewards over a fixed number of design iterations (budget), thereby making the policy adaptive. The choice of a good reward function is essential for informative designs. We discuss various reward function choices, primarily based on an estimate of the true causal graph obtained from a likelihood-free amortized causal discovery approach. Both our policy and the reward function only require access to a simulator of the design environment. Further, we present connections of our approach to amortized sequential Bayesian experimental design~\citep{foster2021deep}. 
We demonstrate that the reward function is related to an approximation of expected information gain based on the amortized posterior distribution over causal graphs. 
As such, CAASL is an intervention design method for performing sample efficient causal structure learning, but is not a new causal structure learning method in itself.

On synthetic data and the single-cell gene expression simulator SERGIO~\citep{dibaeinia2020sergio}, we empirically study various aspects of our trained policy---the amortization performance on training distribution of the design environment as well as on design environments with distribution shifts from the training environment. We find that our policy obtains better causal structure learning performance for a given budget than alternate intervention strategies. 
Overall, we observe excellent generalization capability of the transformer for intervention design, similar to what has been demonstrated in other domains~\citep{brown2020language,kaplan2020scaling,zhai2022scaling}. The robustness of the amortized policy opens up the possibility for lab-in-the-loop intervention design for single-cell data, wherein a single network can propose informative interventions across different cell lines and experimental conditions. 

\section{Background and Related Work}\label{sec:background_related}
\paragraph{Structural Causal Models.} Let $\vy = \{ y_1, \dots, y_d \}$ be the random variables of interest associated with the vertices of a graph $G$. Let $A\in \{0,1\}^{d\times d}$ be the adjacency matrix corresponding to $G$. A Structural Causal Model (SCM)~\citep{peters2017elements} is a framework for causality which consists of a set of equations in which each variable $y_i$ is a deterministic function of its direct causes $y_{\text{pa}_G(i)}$ as well as an exogenous noise variable $\epsilon_i$ with a distribution $P_{\epsilon_i}$
\begin{equation}
\label{eq:scm}
y_i \coloneqq f_i(\vy_{\text{pa}_G(i)}, \epsilon_i; \theta_i).
\end{equation}
The functions $f_i$, with parameters $\theta_i$, are mechanisms that relate how the direct causes affect the variable $y_i$. The structural assignments are typically assumed to be acyclic, with $G$ being a directed acyclic graph whose edges indicate direct causes.
 In addition, an SCM defines the likelihood of any data sample $\vy$ under this model, denoted as $p(\vy\mid \parameters)$. 
 Further, we assume that the SCM is causally sufficient, i.e.~all the variables are measurable (but can be missing at random), and the noise variables are mutually independent. 
 
 \paragraph{Interventions.} The SCM framework admits reasoning about effects of interventions on any variable in $\vy$. Most notable types of intervention include a perfect ($\mathrm{do}$) intervention, and a shift intervention~\citep{rothenhausler2015backshift}. A perfect intervention on any variable $y_i$ corresponds to changing the structural equation of that variable to the desired value, $y_i\coloneqq v_i$. It is denoted by the $\mathrm{do}$-operator~\citep{pearl2009causality} as $\mathrm{do}(y_i=v_i)$. In a shift intervention, the conditional mean of the interventional variable $\E[y_i\mid \vy_{\text{pa}_G(i)}]$ is shifted by $v_i$.  The likelihood of any data under an intervention $I$ is denoted as $p(\vy\mid \parameters,I)$. For perfect and shift interventions, $I$ can be parameterized as a $d\times 2$ dimensional matrix, where the first column corresponds to one-hot encoding of whether a particular variable is intervened or not, and the second column corresponds to the value (or the shift) of the intervention corresponding to each potential intervention target.

\paragraph{Causal Structure Learning.} The problem of causal structure learning corresponds to estimating $A$ (and other parameters of the SCM $\theta$) given samples from $p_\mathrm{data}$~\citep{heinze2018causal}. In general, there could be multiple models (and hence graphs) that can be consistent with a given joint distribution over $\vy$, which necessitates causal structure learning with interventional data~\citep{peters2017elements}. There are various approaches, either based on independence tests~\citep{dai2024gene,spirtes2001causation}, or graph search by maximizing a score function (likelihood of the data with certain assumptions on the SCM)~\citep{brouillard2020differentiable,rolland2022score,hauser2012characterization}. Reinforcement learning has also been used for search over graphs with a score function~\citep{zhu2019causal}, however it differs entirely from our approach wherein we use RL for intervention design. Alternately, based on the tractable likelihood, there are also causal structure learning methods that estimate the posterior distribution $q(A\mid \gD)$ of graphs~\citep{annadani2023bayesdag,deleu2024joint} for a dataset $\gD$ that is sampled from $p_\mathrm{data}$. 

\paragraph{Likelihood-Free Amortized Causal Structure Learning (AVICI)~\citep{lorch2022amortized}.} More recently, instead of inferring causal graphs over specific datasets, amortized posterior inference of causal graphs has also been studied~\citep{lorch2022amortized,ke2022learning}. In particular, the amortized posterior from \citet{lorch2022amortized}, called AVICI, makes use of a transformer to directly predict the posterior $q(A\mid\gD)$ by just a forward-pass of any dataset. The amortized posterior, parameterized as a product of independent Bernoulli random variables over the presence of edges in the causal graph, is trained from a simulator without having access to the likelihood of the data. Since the simulator provides the ground truth value of $A$, the amortized posterior can be trained by maximum likelihood with a combination of observational and interventional data to maximize the probability of the true edges. The AVICI model can amortize over datasets with different dimensionalities $d$, while generalizing well to new datasets that have not been seen during training. Since it is computationally cheap to obtain the posterior distribution with AVICI, we use it for computing the reward for our intervention design policy.

\paragraph{Active Intervention Design.} Active intervention design is the problem of designing interventional experiments to obtain data that enables causal structure learning in a sample efficient manner (under a fixed budget). While adaptive strategies have been explored~\citep{choo2023adaptivity,greenewald2019sample}, these approaches still require intermediate inference of the SCM, and are also not amortized. Intervention design based on Bayesian optimal experimental design~\citep{lindley1956measure,chaloner1995bayesian} has also been considered, although only with additive noise models, which enable likelihood evaluation~\citep{tigas2022interventions,tigas2023differentiable,agrawal2019abcd,zhang2024bayesian, sussex2021near}. Reinforcement learning has also been used in intervention design~\citep{sauter2023meta,lampinen2024passive}, however, they have been limited to non-amortized and small scale settings. 
In contrast to earlier work, we demonstrate the applicability of our method to single-cell simulated gene expression data, wherein the mechanisms are defined by differential equations and also include technical noise. 

\section{Amortized Intervention Design} \label{sec:method}
We first present our active intervention design strategy with reinforcement learning, the corresponding amortized network and its training. In \Cref{sec:eig_connection}, we then present connections of our reward to sequential Bayesian experimental design.

\paragraph{Setting.} Given a budget $T$, intervention design is the problem of finding a sequence of informative interventions with a policy $I_1,\dots,I_T\sim \pi$ that results in an estimate of the causal graph that is close to $A$. For any intervention $I$, a causal model defines a generative model of the data with likelihood $p(\vy\mid \parameters, I)$ and prior $p(A,\theta)$. 
We indicate initial (observational) data, if available, as $\vy_0=\{\vy_0^{(i)}\}_{i=1}^{n_0}$ and the corresponding interventions with $I_0$, where $I_0=\{\emptyset\}^{n_0}$ if the initial data is fully observational. Let $h_t\in \R^{(n_0+t)\times d\times 2}$ denote the interventional history $(\vy_0, I_0),\dots,(\vy_t,I_t)$, obtained by concatenation of $\vy$ and first column of $I$ that correspond to interventional targets. We do not explicitly encode intervention values in history, since for a $\mathrm{do}$ intervention, the intervention values are already present in $\vy$\footnote{We train our policy only on $\mathrm{do}$ interventions.}. Existing intervention design strategies like \citep{tigas2023differentiable} approximate a posterior on $\parameters$ at each step $t$, approximate expected information gain (EIG)~\citep{lindley1956measure,rainforth2024modern} 
and greedily maximize it to compute $I_{t+1}$.
Details of this greedy approach are given in \Cref{sec:eig:greedy}.



\subsection{Intervention Design with Reinforcement Learning}
In this work, we instead treat intervention design as a Reinforcement Learning (RL) problem and train a single policy network $\pi_\phi$ with parameters $\phi$ to obtain a sequence of adaptive interventions $I_1,\dots,I_T$ for any underlying causal graph with adjacency matrix $A$. In order to do so, we first describe the RL environment under which the interventions are performed.

\paragraph{Intervention Design Environment.} Similar to \citet{blau2022optimizing}, we define an interventional design environment as a Hidden-Parameter Markov Decision Process (HiP-MDP)~\citep{doshi2016hidden}. The HiP-MDP  we use, $\gM(\parameters)$, has hidden parameters $\parameters$ and can be fully described by the tuple $(\gS, \gA, \rho, \beta, \gT, R, \gamma, \, p_\beta)$. The state-space $\gS$ consists of the histories $s_t=h_t$, the initial state $\rho=h_0=(\vy_0, I_0)$ corresponds to initial data, the action-space $\gA$ corresponds to interventions $a_t=I_t$ and $\beta$ describes the space of all causal models (graphs and parameters) with prior $p_\beta=p(A,\theta)$.
The hidden parameters are sampled for each episode at the beginning from the prior. $\gamma$ is the discount factor. In a HiP-MDP, the transition function $\gT$ and reward $R$ depend on the hidden parameters. The transition function $\gT(h_t\mid h_{t-1}, I_t, \parameters)$ is Markovian, and it involves two operations: (1) sampling interventional data $\vy_t\sim p(\vy\mid \parameters, I_t)$, and (2) updating the history state $h_t=\mathrm{Concat}[h_{t-1},(\vy_t,I_t)]$. For a reward function $R(h_t, I_t, h_{t-1},\parameters)$ that we define below, intervention design corresponds to finding the parameters $\phi$ of the amortized policy that maximizes the expected cumulative reward of all interventions:
\begin{align}\label{eq:cum_reward}
    \max_{\phi} \E_{\pi_\phi,\rho,\gT, \,p(A,\theta)}&\left[\sum_{t=1}^T \gamma^{t-1} R\left(h_t, I_t, h_{t-1},\parameters\right)\right]\\
     \text{with}\,\,\,\,&I_t\sim \pi_\phi(h_{t-1})\nonumber
\end{align}

\paragraph{Reward Function.} For the purpose of amortized intervention design, a good reward function should be cheap to evaluate while leading to informative interventions. 
In this work, we propose to utilize the estimate of the causal graph from an amortized causal graph posterior $q(\hat{A}\mid h_{t})$. In particular, we use the pretrained AVICI model~\citep{lorch2022amortized}. AVICI is a transformer based neural network trained with (interventional) data from a simulator to directly predict the probability of presence or absence of any edge in the causal graph by just a forward pass of the data, without requiring access to the likelihood. For any history $h_{t-1}$, we define the reward for performing intervention $I_t$ and reaching state $h_t$ as the improvement in the number of correct entries in the predicted adjacency matrix of the AVICI model:
\begin{equation}\label{eq:reward}
    R(h_t, I_t, h_{t-1}, \parameters) = \E_{q(\hat{A}\mid h_t)}\left[\sum_{i,j}\sI\left[\hat{A}_{i,j}={A}_{i,j}\right]\right] - R(h_{t-1}, I_{t-1}, h_{t-2}, \parameters)
\end{equation}
where $\sI[\cdot]$ is the indicator function and $R(h_0, I_0, \parameters)
=\E_{q(\hat{A}\mid h_0)}\left[\sum_{i,j}\sI\left[\hat{A}_{i,j}={A}_{i,j}\right]\right]$. 
We note that our choice of reward function revolves around obtaining a good estimate for the causal graph, $A$; we do not (directly) reward learning about $\theta$. 

The above RL problem for intervention design is intuitive: reward the intervention in proportion to the improvement it brings in terms of number of correct entries of the adjacency matrix from the amortized posterior. Also, for any $t$, the cumulative reward, \cref{eq:cum_reward}, for $\gamma=1$ of all interventions including $I_0$, which includes an additional term $R(h_0, I_0, \parameters)$, is simply the number of correct entries of the adjacency matrix predicted by the amortized posterior model for $h_t$.
This reward telescoping was inspired by \citet{blau2022optimizing}.
We also show in \Cref{sec:eig_connection} that this reward function is also related to an approximation of multi-step EIG, the quantity of interest in sequential Bayesian experimental design~\citep{foster2021deep}.
\subsection{Policy} 

\paragraph{Architecture.} In order for the policy to achieve amortization and generalize to new environments not seen during training, it should encode key design space symmetries. In particular, for the problem of intervention design, the interventions should be permutation equivariant to ordering of the variables and permutation invariant to the ordering of the history. This can be ensured by a transformer architecture~\citep{vaswani2017attention} wherein self-attention is applied alternately---once over the variable axis and next over the samples axis~\citep{kossen2021self}. More precisely, we input $h_t\in \R^{(n_0+t)\times d\times 2}$ and apply self-attention\footnote{Every self-attention is multi-headed, followed by a position-wise feedforward layer and layer normalization, as in a standard transformer block.} over first the $n_0+t$ axis and next over the $d$ axis. This ensures that the history representation is permutation equivariant over both the axes~\citep{lee2019set}. After multiple layers of alternating self-attention, we apply max pooling over the samples (dim.~$n_0+t$) axis, which gives an encoding of size $l$ of the history $B_t\in\R^{d\times l}$ that respects the desired symmetries. The same symmetries apply for amortized causal structure learning, hence the reward model AVICI also leverages the alternate attention architecture.
The history embedding $B_t$ is then passed through a multi-layer perceptron, whose outputs parameterize the logits of the Gaussian-Tanh distribution~\citep{haarnoja2018soft}, from which the interventions are sampled. In our setting, we model both the intervention targets and intervention values, hence $I_t$ is $d \times 2$ dimensional. Gaussian-Tanh samples range from -1 to 1. We use $I_t[:, 0]$ to encode the interventional targets by discretizing the values to a binary mask (0 and 1) by thresholding at 0, where 1 indicates intervention on the variable $y_i$. If an intervention on $y_i$ is active, the value to intervene with is given by $I_t[i, 1]$.

\paragraph{Training.} Training the policy involves addressing two main challenges: computing the reward in \cref{eq:reward} since $\parameters$ would be unknown for real datasets, and optimizing this reward, which is discrete. 
In order to address the first challenge, we simulate interventional data $\vy_t\sim p(\vy\mid \{A',\theta'\}, I_t)$ for a sample $\{A',\theta'\}\sim p(A,\theta)$ from the prior using a \emph{simulator}. Such simulators exist for single cell gene regulatory networks (e.g. \citep{dibaeinia2020sergio,cannoodt2021spearheading,schaffter2011genenetweaver}) and are becoming increasingly widespread in other domains~\citep{gamella2024causal,ahmed2020causalworld}. The reward model AVICI is pretrained on datasets from the prior $p(\parameters)$ using the same simulator. During training of the policy, we only use the pretrained reward model for inference and do not update its parameters. To address the second challenge, we train our policy using Soft-Actor Critic (SAC)~\citep{haarnoja2018soft}, an off-policy reinforcement learning algorithm that does not require rewards to be differentiable. We use the REDQ version of SAC to improve sample efficiency~\citep{chen2021randomized}. REDQ trains multiple Q-function networks to optimize the reward. For each Q-function network, we use a transformer based history state encoder with architecture similar to that in the policy, but the weights are not shared. This is beneficial because the same equivariance--invariance properties that hold for the policy should also hold for the Q-function.

\paragraph{Inference.} 
Deploying the policy in a real (i.e.~not simulated) environment 
amounts to a rollout of the policy through interaction with the real environment. This requires just a forward pass of policy network for each time step $t$.
Note that we do not need intermediate Bayesian inference or other estimation of the causal graph on the collected data. 


\section{Choice of Reward Function}
\subsection{Connection to Sequential Bayesian Experimental Design}
\label{sec:eig_connection}

As discussed in \Cref{sec:eig}, the problem we tackle has a connection to likelihood-free sequential Bayesian experimental design \citep{foster2021deep,ivanova2021implicit}.
With the aim of gathering data to learn about the causal graph $A$, the multi-step expected information gain (EIG) can be written
\begin{align}\label{eq:A_eig_main}
    \text{EIG}(A;\pi_\phi) = \E_{\pi_\phi,\rho,\gT, \,p(\parameters)}\left[\log p(A \mid h_t) \right] + \text{const}.
\end{align}
Since the posterior $p(A \mid h_t)$ is intractable, we could replace it by an approximate posterior $q(A \mid h_t)$. This gives rise to the Barber-Agakov (BA) bound \citep{barber2004algorithm,foster2020unified}, which was recently explored in a sequential context by \citet{blau2023cross}.
This tells us that we have an EIG lower bound by using $q$ in place of $p$:
\begin{equation}\label{eq:ba}
    \text{EIG}(A;\pi_\phi) \ge \E_{\pi_\phi,\rho,\gT, \,p(\parameters)}\left[\log q(A \mid h_t) \right] + \text{const}.
\end{equation}
We can interpret \cref{eq:ba} in simple terms---taking $\log q(A \mid h_t)$ as a reward function is equivalent to optimizing a lower bound on the EIG.
Although \cref{eq:ba} implies that we only receive a reward on the final state $h_t$, it is possible to rewrite this using telescoping rewards \citep{blau2022optimizing} exactly as we do in  \cref{eq:reward}.
The BA bound therefore represents the closest point of comparison between the method we outline in \Cref{sec:method} and sequential Bayesian experimental design.
As with the BA bound, we make use of an amortized approximate posterior distribution $q(A \mid h_t)$ that works backwards from data $h_t$ to predict the graph that might have generated it. 
Unlike the BA bound, however, we use the adjacency matrix accuracy to compare the true $A$ to samples $\hat{A}$ from the amortized posterior, rather than computing the log-likelihood of the true graph under that amortized posterior, $\log q(A \mid h_t)$.
We found that this worked better in practice.
Nevertheless, we see a close relationship between the approach we take and the methods of sequential Bayesian experimental design.



\subsection{Other Possible Reward Functions}

Any target metric for causal structure learning like structural hamming distance computed on the amortized posterior could be used as a reward function. Depending on the application, domain specific causal graph objectives could also be considered. While a large number of possibilities exist, we use expected number of correct entries of adjacency matrix, \cref{eq:reward}, as the reward for training CAASL. As opposed to Structural Hamming Distance (SHD) and Area Under Precision Recall Curve (AUPRC), \cref{eq:reward} is straightforward to compute and parallelize. 
\begin{figure}
    \centering
    \includegraphics[width=0.85\linewidth]{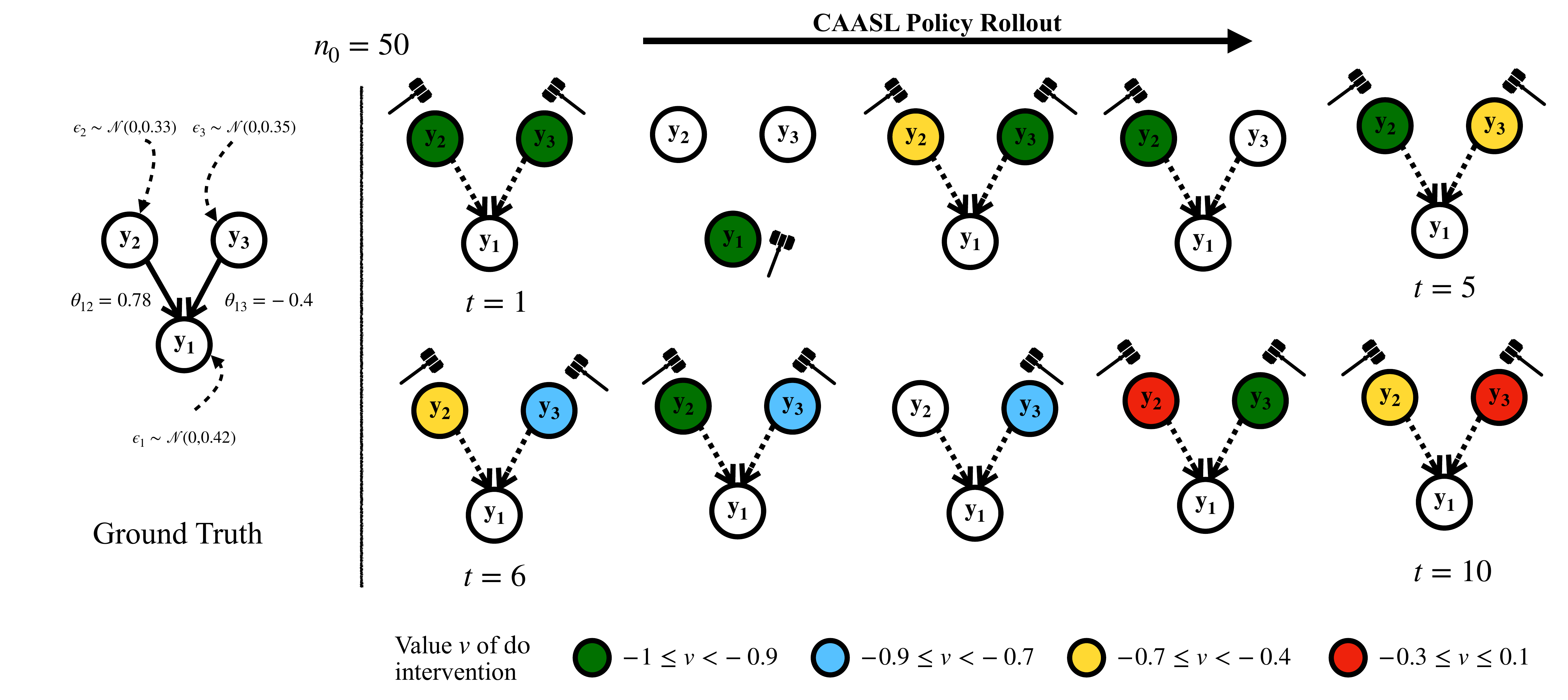}
    \caption{Visualization of the rollout of the trained CAASL policy on a randomly sampled environment with $n_0=50$ initial observational samples. Colored circles indicate nodes with a $\mathrm{do}$ intervention. The policy selects interventions that mostly correspond to the variables with a child in the ground truth graph. At $t=2$, the policy selects the only child $y_1$, which breaks all direct causal effects. This gives lesser information about the overall causal model. After this,  $y_1$ is never chosen. Initially, the policy is exploratory wrt targets and exploitative wrt values. This trend is reversed as the episode progresses. The policy is trained on environments with $d=2$, therefore it has not seen any graphs with $d=3$ before.}
    \label{fig:policy_viz}
\end{figure}

\section{Experiments}\label{sec:exp}
We train CAASL policy on two challenging environment domains: 1. Synthetic design environment with a causal model defined by linear mechanisms and additive noise, and 2. SERGIO~\citep{dibaeinia2020sergio}, a single-cell simulator corresponding to any gene regulatory network.  For each domain, we train a single CAASL policy on a distribution of design environments with $d=10$. A distribution of intervention design environments is defined by the choice of prior over causal models $p(A,\theta)$, which includes priors over graphs $A$ (e.g. Erdős–Rényi~\citep{erdHos1960evolution}),  mechanism parameters $\theta$ and noise. We define an Out-of-Distribution (OOD) environment as any environment with the choice of prior (either over graphs, mechanisms parameters or noise) that is different from training. In addition to these distribution shifts, we also consider OOD environments wherein the priors remain the same, but either the dimensionality of the data $d$ increases (i.e. $d>10$), or the performed intervention type changes. Precise choice of training and OOD testing distributions are given in \Cref{app:env_train_ood}. All evaluation experiments are conducted \textit{in silico}, on environments with causal model parameters that CAASL has never seen during
training, regardless of whether the environment is in-distribution or OOD. In addition, all evaluation is done by just forward passing the history through the policy.

\paragraph{Baselines.} We compare our approach with two amortized strategies: Random and Observational. Random corresponds to obtaining data from random interventions, while Observational corresponds to collecting more observational data. For the synthetic design environment domain, we also compare with DiffCBED~\citep{tigas2023differentiable} and SS Finite~\citep{sussex2021near}. These intervention strategies use likelihood of the data to perform designs. So in certain OOD synthetic design environments and the single-cell simulator SERGIO where the likelihood is not available, we omit these baselines. DiffCBED and SS Finite rely on an approximate causal graph posterior to design interventions. As suggested in \citep{tigas2023differentiable}, we use bootstrapped GIES~\citep{hauser2012characterization,friedman2013data} as the approximate posterior distribution for these baselines. For an evaluation task on 100 random design environments with a budget of 10, DiffCBED and SS Finite methods require approximate posterior inference of the causal graph 1000 times.

\paragraph{Metrics.} All evaluation is done on 100 random test environments. As CAASL is an intervention design method, we measure the cumulative rewards with $\gamma=1$ (returns) obtained from the graph predicted by the amortized posterior. However, for the sake of completeness, we also measure structure learning related metrics like the Structural Hamming Distance (SHD), the Area under Precision Recall Curve (AUPRC) and F1 score (Edge F1) between the graph predicted by the amortized posterior and the true graph~\citep{lorch2022amortized,annadani2023bayesdag}. Precise definition of these metrics is provided in \Cref{app:all_res}. We find that in most cases all the metrics are correlated. Therefore, unless otherwise mentioned, we only report the returns and relegate the other metrics to \Cref{app:all_res}.

\begin{figure}
    \centering
    \includegraphics[width=\linewidth]{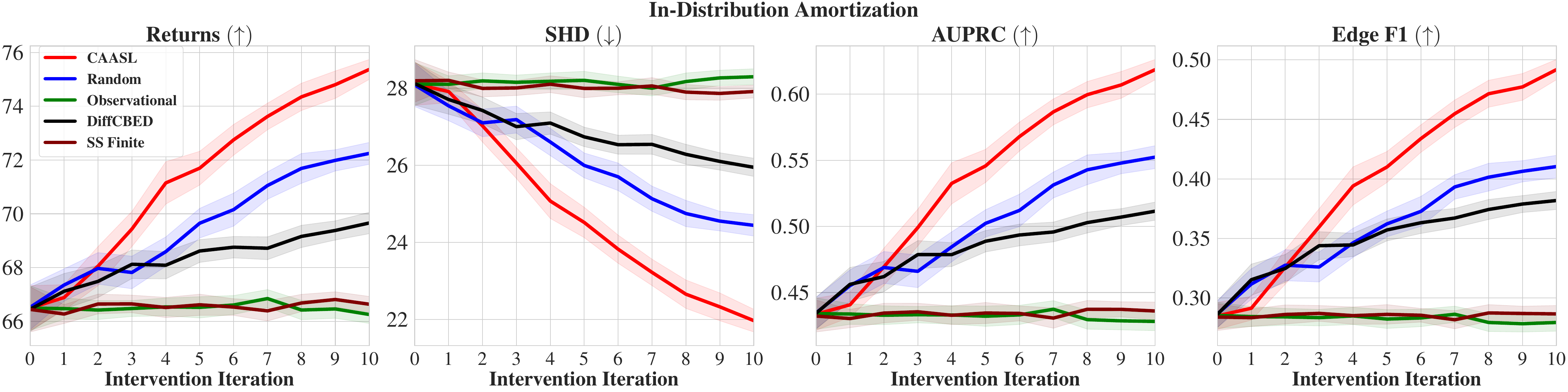}
    \caption{Amortization results of various intervention strategies on 100 random test environments. CAASL significantly outperforms other intervention strategies. Shaded area represents 95\% CI. }
    \label{fig:lin_anm_agg}
\end{figure}
\subsection{Synthetic Design Environment}\label{sec:exp_synthetic}
\paragraph{Training Distribution of the Design Environment.} We train CAASL on synthetically generated data, wherein $p(A, \theta)$ consists of linear SCMs with additive homoskedastic Gaussian noise. The dimensionality during training is $d=10$. The prior over causal graphs is Erdős–Rényi~\citep{erdHos1960evolution}, with 3 edges per node in expectation. The prior over linear coefficients is chosen such that the marginal variance of each variable is close to $1$. This is done to ensure that structure learning algorithms are not sensitive to the scale of the data~\citep{reisach2021beware}. During training, an intervention exclusively corresponds to a $\mathrm{do}$ intervention. Further, we set $n_0=50$ and the budget $T$ is fixed to $10$.

\paragraph{Training Details.} We train CAASL with $4$ layers of alternating attention for the transformer, followed by a max pooling operation over the history, to give an embedding with size $l=32$. SAC related hyperparameters are tuned based on performance on held-out design environments. Details of the architecture, hyperparameter tuning and optimizer is given in \Cref{app:training_details}. For the reward model, we use AVICI that is pretrained on random linear additive noise datasets.

\paragraph{Amortization Performance.} We test on novel environments with hidden parameters sampled from the training prior $p(A,\theta)$. Results are provided in \cref{fig:lin_anm_agg}. We find that our policy significantly outperforms the random baseline in terms of returns as well as more common structure learning metrics like the SHD, AUPRC and Edge F1. For instance, our method achieves returns close to $76$ with just $10$ interventional samples, while the random baseline achieves close to 72. Other intervention strategies like DiffCBED~\citep{tigas2023differentiable} and SS Finite~\citep{sussex2021near} perform worse, while still making use of the likelihood and performing intermediate inference of causal structure.
\begin{figure}
    \centering
    \includegraphics[width=\linewidth]{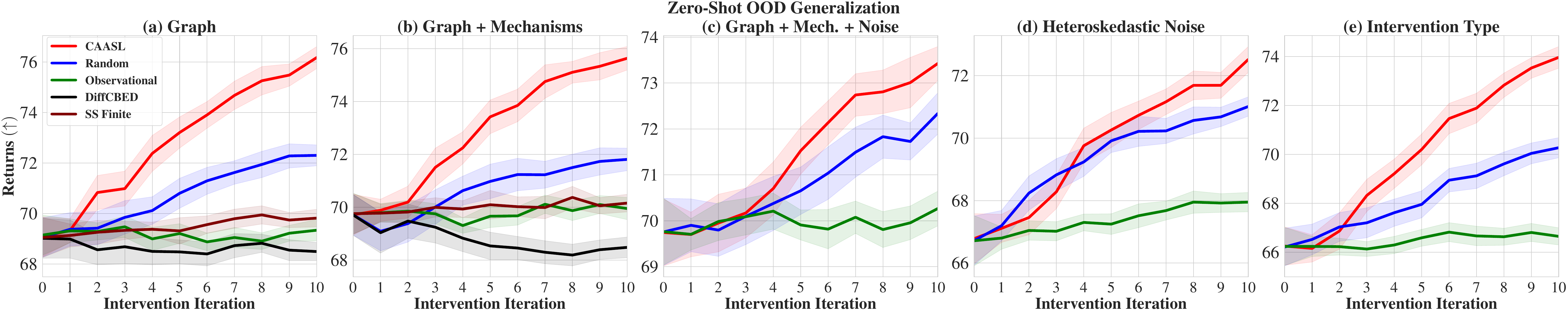}
    \caption{\looseness=-1 Zero-shot OOD returns of CAASL on 100 random environments with distribution shift coming from (a) graphs, (b) graphs and mechanisms, (c) graphs, mechanisms and noise, (d) noise changes from homoskedastic to heteroskedastic, and finally (e) intervention changes from $\mathrm{do}$ to a shift intervention. CAASL outperforms other intervention strategies. Shaded area represents 95\% CI.}
    \label{fig:lin_anm_ood}
\end{figure}

\paragraph{Zero-Shot OOD Generalization.} We also test the trained CAASL policy on environments when the prior changes. All results correspond to zero-shot performance, obtained by just a forward forward pass of the trained policy. \cref{fig:lin_anm_ood} presents the returns of CAASL alongside other applicable baselines. We consider shifts which become increasingly different from training: (1) the graph prior changes from Erdős–Rényi~\citep{erdHos1960evolution} to Scale-Free~\citep{barabasi1999emergence} (\cref{fig:lin_anm_ood} (a)), (2) apart from the graph, prior over mechanisms also change (\cref{fig:lin_anm_ood} (b)), (3) apart from graph and mechanisms, the noise distribution changes from Gaussian to Gumbel (\cref{fig:lin_anm_ood} (c)). We find that our policy achieves better performance than random strategy by a significant margin. Further, our method also outperforms DiffCBED and SS Finite which explicitly optimize for designs corresponding to these environments. In addition to these OOD settings, we also consider OOD environments in which the prior remains the same, but the noise is heteroskedastic instead of homoskedastic (\cref{fig:lin_anm_ood} (d)). Although the random strategy is very competitive, CAASL performs better. Finally, we consider OOD environments wherein the intervention design suggested by the policy during testing is used for performing a shift intervention instead of a $\mathrm{do}$ (\cref{fig:lin_anm_ood} (d)). CAASL performs better than baselines even in this setting. 
\begin{wrapfigure}{r}{0.5\linewidth}
    \vspace{-0.1cm}
    \includegraphics[width=0.95\linewidth]{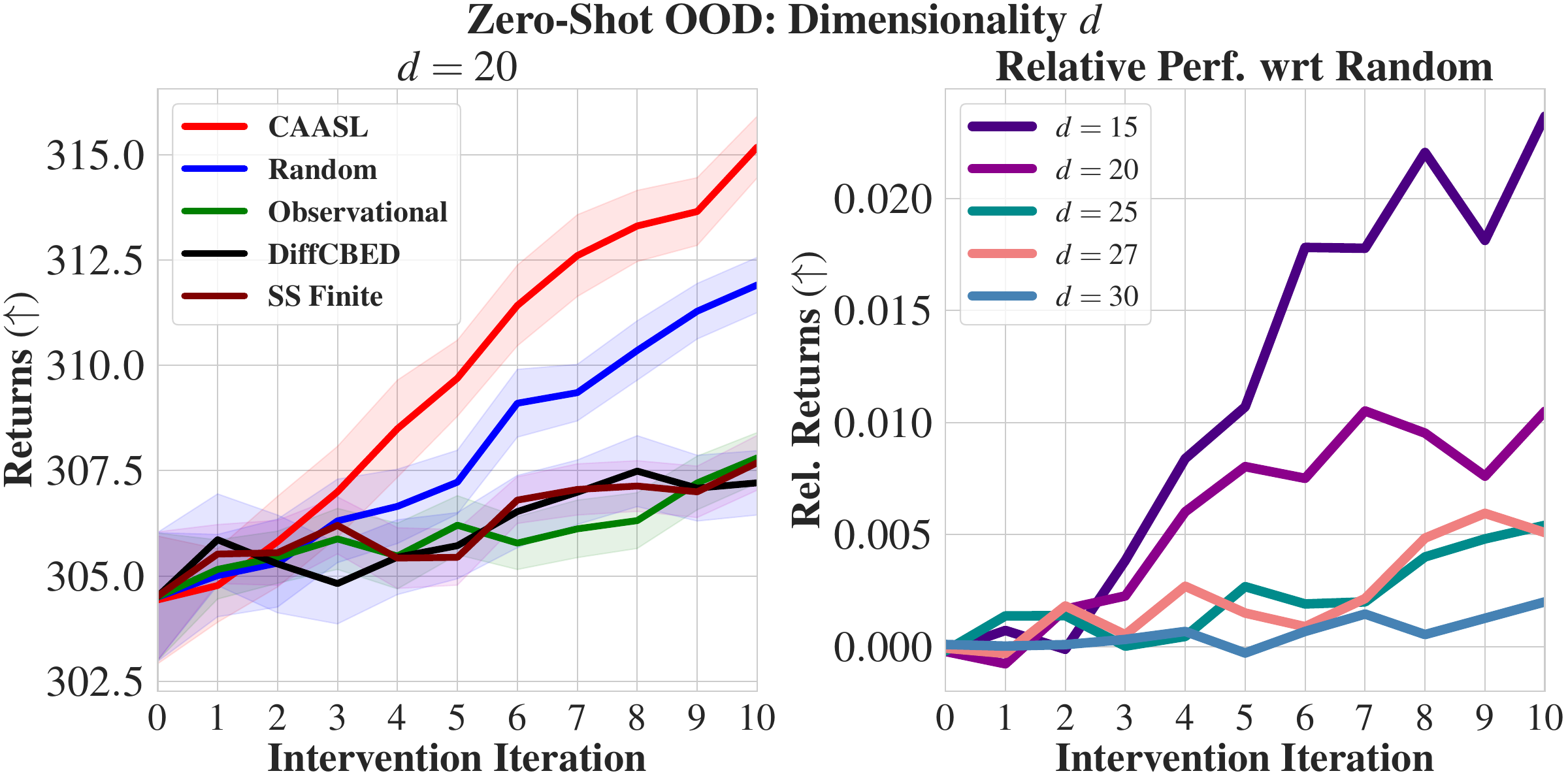}
    \caption{Zero-Shot OOD generalization results when dimensionality $d$ changes for synthetic environment. For training, $d=10$. Left: Zero-Shot test returns with $d=20$. Right: Relative mean zero-shot returns of CAASL wrt random for different $d$. Results on 100 random environments. Shaded area represents 95\% CI.}
    \label{fig:lin_ood_d}
     \vspace{-1.3cm}
\end{wrapfigure}

Slightly different to the above OOD environments, we also consider OOD environments in which the dimensionality of the data changes during testing, but the prior remains the same. \cref{fig:lin_ood_d} presents the results, with further details in \cref{fig:linanm_multiple_d}. CAASL obtains better returns on average than random at all points of acquisition. The relative performance of CAASL decreases as $d$ increases (up to $d=30$) from training, although it still performs better than random. 

\subsection{Single-Cell Gene Regulatory Network Environment}\label{sec:exp_sergio}
In this setting, we train a CAASL policy based on the single-cell gene expression simulator SERGIO~\citep{dibaeinia2020sergio}. Given a causal graph that corresponds to interaction between different genes in terms of their transcription regulation, SERGIO simulates expressions of genes that correspond to steady state of differential equations that govern the interaction between the genes. Each variable entry indicates the count of mRNA that is produced corresponding to that gene, similar to the output of modern single-cell RNA sequencing (scRNA-seq) technological platforms~\citep{macosko2015highly}. In addition, SERGIO can be extended to support interventions. Interventions in this setting correspond to either gene knockouts, wherein the transcription rate of the intervened gene is actively set to 0, or gene knockdown, wherein the intervened gene's transcription rate is actively halved. Since there is no value selection in this setting, the dimensionality of the policy is $d$ instead of $d\times 2$. SERGIO also simulates technical noise such that the statistics of the data match that obtained from real scRNA-seq platforms. Some of the technical noise includes dropouts (missingness of the data), library size effects and random outlier effects. Most notably, atleast $70\%$ of the data is missing in most single cell platforms. Therefore, in this domain, not only is the likelihood intractable, but also there is high amount of missing data. We do not impute the missing data, but just encode it with $0$. 

\paragraph{Training Distribution of the Design Environment.} For training, we set $d=10$, with $n_0=50$ observational (wild-type) data with budget $T=10$. The statistics of the data corresponds to 10X Chromium platform~\citep{dibaeinia2020sergio} wherein around 74\% of the data is dropped out. The prior over causal graphs is set to Erdős–Rényi~\citep{erdHos1960evolution} with 3 edges per node on average. An intervention exclusively corresponds to a gene knockout. We provide details of the simulator in \cref{app:sergio_details} and the training prior parameters in \cref{app:sergio_train_ood}. 

\paragraph{Training Details.} We train CAASL with 3 layers of alternate attention, followed by a max pooling operation, giving an embedding of size $l=32$. Just like in the synthetic linear domain, SAC related hyperparameters are tuned based on performance on held-out design environments. Details are given in \cref{app:training_details}. Once trained, we perform a forward pass of the history through the policy to obtain intervention designs for all test environments. For the reward model, we use AVICI that is pretrained on this simulator with post-noise data statistics matching that of 10X chromium platform.
\begin{figure}
    \centering
    \includegraphics[width=\linewidth]{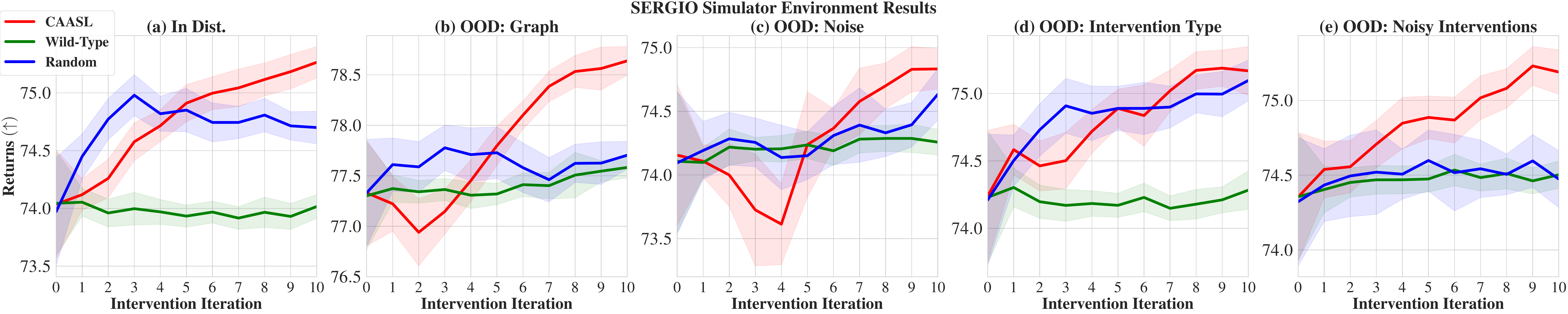}
    \caption{Results on SERGIO environment with 100 random environments. (a) corresponds to in-distribution performance, (b)-(e) correspond to zero-shot OOD performance with distribution shift coming from either (b) graphs, (c) technical noise, (d) intervention changing to a gene-knockdown (e) Noisy interventions, which include off-target effects. Shaded area represents 95\% CI.}
    \label{fig:results_sergio}
\end{figure}
\paragraph{Amortization Performance.} The in-distribution amortization performance is presented in \cref{fig:results_sergio}(a). After 5 acquisitions, CAASL obtains better returns than random.  

\paragraph{Zero-Shot OOD Generalization.} We test the CAASL policy when the environment is subject to various test-time distribution shifts. Robustness to distribution shifts is important in real world-settings, where experimental conditions can change. We consider 4 different OOD environments: (1) the prior over graphs changes from Erdős–Rényi to Scale-free (\cref{fig:results_sergio}(b)), (2) The perturbation platform changes to Drop-Seq~\citep{macosko2015highly}, wherein among other noise parameters, the amount of missing data increases from 74 to 85\% (\cref{fig:results_sergio}(c)), (3) The intervention type changes from knockout to knockdown (\cref{fig:results_sergio}(d)) and, (4) Noisy knockout interventions, where there is a 10\% chance that either the intended gene does not get knocked out, or an off-target gene is knocked out (\cref{fig:results_sergio}(e)). We find that CAASL shows excellent robustness to these distribution shifts, and obtains better returns than baselines. When the intervention type changes, the random baseline is still competitive. An interesting observation is that the for the OOD graph and the OOD noise setting, the model shows exploratory behavior in the beginning where the returns decrease, but later becomes better than random. Robustness to various distribution-shifts demonstrates the generality of the policy.
\paragraph{Limitation.} For the zero-shot OOD generalization when the dimensionality of the data increases, we noticed that the performance of CAASL is on par with random, but is not necessarily better (\cref{fig:sergio_d}). We hypothesize that since almost $74\%$ of data is missing, the incorporated design space symmetries might not be as relevant, which might limit the extent of zero-shot generalization.

\section{Conclusion}
 We have presented an amortized and adaptive intervention design strategy CAASL, that does not require intermediate inference of the causal graph. CAASL is based on a policy parameterized by the transformer which is permutation equivariant to ordering of the variables and permutation invariant to ordering of the collected data. Through various experiments, including on a simulator which respects the data statistics of real gene-expression readouts, we find that our method shows excellent amortized intervention design and zero-shot generalization to significant distribution shifts.  The achieved performance motivates intervention design in more complex settings -  high-throughput experiments with large batch sizes and utilization of existing real offline data for designing interventions.

\section*{Acknowledgements}
The authors gratefully acknowledge the Gauss Centre for Supercomputing e.V. {\url{www.gauss-centre.eu}) for funding this project by providing computing time on the GCS Supercomputer JUWELS\citep{JUWELS} at Jülich Supercomputing Centre (JSC).

\bibliographystyle{plainnat}
\bibliography{references.bib}

\appendix

\section{Connections to Bayesian Experimental Design using Expected Information Gain} \label{sec:eig}
\subsection{Greedy Approaches} \label{sec:eig:greedy}
We consider the model with unknown parameters $\parameters$, prior $p(\parameters)$ and likelihood of the data $p(\vy\mid \parameters, I)$ under an intervention $I$. The Expected Information Gain (EIG) is given by:
\begin{equation}\label{eq:EIG}
    \mathrm{EIG}(I)= \E_{p(\parameters)p(\vy\mid \parameters, I)}\left[\log p(\vy\mid \parameters, I)-\log p(\vy\mid I)\right].
\end{equation}
In the standard greedy approach to Bayesian experimental design \citep{rainforth2024modern}, given a history $h_{t-1}$, we replace the prior $p(\parameters)$ with the posterior conditional on existing data $p(\parameters \mid h_{t-1})$ and then estimate the one-step EIG
\begin{equation}\label{eq:EIG_step}
    \mathrm{EIG}(I)= \E_{p(\parameters \mid h_{t-1})p(\vy\mid \parameters, I)}\left[\log p(\vy\mid \parameters, I)-\log p(\vy\mid h_{t-1}, I)\right].
\end{equation}
where $p(\vy\mid h_{t-1}, I) = \int_{A,\theta} p(\parameters \mid h_{t-1})p(\vy\mid \parameters, I)$.
The EIG is estimated for each candidate design $I$, and the one with the largest EIG is selected.
This gives rise to the policy $\pi_\mathrm{greedy}$, which was applied to causal graph discovery by e.g.~\citet{tigas2023differentiable}.

\subsection{Non-greedy Approaches}
Non-greedy approaches to experimental design using EIG were also explored \citep{huan2016sequential,foster2021deep}. Using the parameters $\parameters$ of our model and the notation of \citet{foster2021deep}, the EIG of a sequence of $t$ experiments generated using policy $\pi_\phi$ about $\parameters$ is given by
\begin{align}\label{eq:seqEIG}
    \mathrm{EIG}(\parameters; \pi_\phi)&= \E_{p(\parameters)p(h_t\mid \parameters, \pi_\phi)}\left[\log p(h_t\mid \parameters,\pi_\phi)-\log p(h_t \mid \pi_\phi)\right] \\
    \text{where} & \ p(h_t \mid \parameters,\pi_\phi) = \prod_{\tau=1}^t p(I_\tau \mid \pi_\phi(h_{\tau-1}))p(\vy_\tau \mid \parameters, I_\tau)
\end{align}
and $p(h_t \mid \pi_\phi)$ is the marginal of this quantity over $p(\parameters)$.
\Cref{eq:seqEIG} cannot be computed exactly, so likelihood-based \citep{foster2021deep} and likelihood-free \citep{ivanova2021implicit} approximations have both been explored.
The likelihood-based sPCE lower bound on EIG \citep{foster2021deep} was also used as a reward function to train an RL policy \citep{blau2022optimizing}.

The problem we consider in this paper is decidedly likelihood-free for two reasons: (1) some simulators do not have explicit likelihoods, (2) even where a likelihood is available, it is generally conditional on both $A$ and $\theta$. We made the choice to focus on experimental design to learn $A$ (ignoring information gain about $\theta$). In this case, the relevant EIG is
\begin{align}\label{eq:eig:A}
    \mathrm{EIG}(A; \pi_\phi)&= \E_{p(\parameters)p(h_t\mid \parameters, \pi_\phi)}\left[\log p(h_t\mid A,\pi_\phi)-\log p(h_t \mid \pi_\phi)\right].
\end{align}
We would have to perform a costly marginalization over $\theta$ to obtain the relevant likelihood, $p(h_t \mid A,\pi_\phi)$.

\Cref{eq:eig:A} can be rearranged using Bayes Theorem to read
\begin{align}
    \mathrm{EIG}(A; \pi_\phi)&= \E_{p(\parameters)p(h_t\mid \parameters, \pi_\phi)}\left[\log p(A \mid h_t)-\log p(A)\right] \\ &= \E_{p(\parameters)p(h_t\mid \parameters, \pi_\phi)}\left[\log p(A \mid h_t)\right] + \text{const}, \label{eq:eig:A_posterior}
\end{align}
where we make the observation that $\E[-\log p(A)]$ is a constant with respect to the design policy $\pi_\phi$.
The form of EIG \cref{eq:eig:A_posterior} is the jumping off point for the BA bound, in which we replace $p(A \mid h_t)$ with an approximate posterior.

\section{Details of Design Environments}
\subsection{Synthetic Design Environment}\label{app:synth_details}
We consider linear additive noise models. For homoskedastic noise, they can be written as
\begin{equation}\label{eq:anm}
    y_i \coloneqq \theta_i^T \vy_{\textbf{pa}_G(i)} + \epsilon_i
\end{equation}
where $\theta_i\sim p(\theta)$ and $\epsilon_i\sim p_\mathrm{noise}$ which can be either a Gaussian or a gumbel distribution.

For heteroskedastic noise, the above equation can be written as:
\begin{equation}\label{eq:anm_hk}
    y_i \coloneqq \theta_i^T \vy_{\textbf{pa}_G(i)} + \sigma_i(\vy_{\textbf{pa}_G(i)})\cdot\epsilon_i
\end{equation}
where $\sigma_i(\cdot)$ is scaling factor that is obtained by a squash operation $\sigma_i(\vx)=\log(1+\exp(g_i(\vx)))$ on any nonlinear function $g_i$. Similar to \citep{lorch2022amortized}, we implement $g_i$ with 100 random Fourier feature features~\cite{rahimi2007random}. Random Fourier feature functions require a kernel, for which we use a Squared-Exponential Kernel with length $ls=10$ and output scale $os=2$.
\subsection{Single-Cell Gene Regulatory Network Environment}\label{app:sergio_details}
SERGIO~\citep{dibaeinia2020sergio} is a single-cell simulator of gene expression for any user provided gene-regulatory network that resembles the data obtained with modern single-cell RNA sequencing (scRNA-seq) technologies technologies like Drop-Seq~\citep{macosko2015highly} and 10X Chromium~\citep{dibaeinia2020sergio}. 

We provide a brief overview of simulation procedure of SERGIO. Our simulation is based on the original simulator provided by \citet{dibaeinia2020sergio} and hence further details can be found in the paper. This simulator was extended by \citet{lorch2022amortized} to support knockouts and knockdowns. We further vectorize the simulator to produce datasets for multiple regulatory networks parallely. The simulator of \citet{dibaeinia2020sergio} is publicly available under GPL-3.0 license.

\subsubsection{Simulation of Gene Expressions}
We first describe the data that is generated without interventions, also called as wild-type measurements. Later, we then describe interventional simulations.

For an observational dataset of size $n\times d$, the data produced from the simulator corresponds to the count of mRNA corresponding to each gene of $n$ different single cells. In particular, given a regulatory network $A$, steady-state of the regulatory differential equation is simulated for $n$ single cells that is regulated according to $A$. SERGIO allows for biological variations within this pool of $n$ single cells, such as varying basal rates of master regulator genes. A master regulator gene is a gene with no upstream genes in $A$. The transcription rate of a master regulator gene is usually a constant, called the basal rate. Usually, cell of the same \textit{type} have the same basal rates. For simulation, we consider $c$ cell types and $n_c$ single cells of each type such that $n = c\cdot n_c$. In this work, we fix $c=5$. If $n$ is less than 5, we sample $n$ single-cells at random after simulating $5$ single-cells corresponding to each cell type. The expression of all the downstream genes is effected nonlinearly by the mRNA production and decay of their respective regulatory genes. This expression is simulated according to a Langevin equation. Finally, the clean data is the continuous-valued mRNA concentration that is measured at random-time points of the steady-state Langevin simulation.

The clean data is then subject to technical measurement noise. The series of simulated noise is as follows:
\begin{enumerate}
    \item With probability $p_\mathrm{outlier}\in [0,1]$, a gene is converted to an outlier gene that has unusually high expression across different cells. This is done by multiplying the current expression with values from a log-normal with mean $\mu_\mathrm{outlier}$ and scale $\sigma_\mathrm{outlier}$.
    \item Based on the single-cell pool considered, different cells have different count distribution data. This is called as library-size effect, which is modeled as a log-normal distribution with mean $\mu_\mathrm{lib}$ and scale $\sigma_\mathrm{lib}$.
    \item The dropouts are simulated with parameters dropout percentile $\delta\in[0,100]$ and the temperature of the logistic function $\eta\in \R_+$
\end{enumerate}
The actual values of the noise parameters differs across different scRNA-seq technologies. The final scRNA-seq resembling simulated data is obtained by sampling from a poisson distribution that is parameterized by the post-noise mRNA concentration levels. 

During a gene knockout, the upstream genes do not effect the knocked out gene. The activity of the gene is set to 0 and is propagated downstream as before. In gene knockdowns, the upstream genes still work the same way as before, however, the expression of the knocked down gene is multiplied by 0.5 for every time step of the steady state simulation. The reduced gene expression of the knocked down gene is propagated to downstream genes as before.
\subsubsection{Simulation Parameters}\label{app:sergio_sim_params}
For generating the clean data, we use the following parameters across all settings, which is similar to what is used for training the reward model~\citep{lorch2022amortized}:
\begin{itemize}
\item Number of cell types $c=5$.
    \item Basal rates $b\sim \mathrm{Uniform}(1,3)$.
    \item Rate of decay of each gene $\lambda=0.8$.
    \item Langevin equation related parameters: Hill function coefficient $\gamma=1$, system noise scale $\epsilon_s=1.0$, interaction strength $k\sim \mathrm{Uniform}(1,5)$ and the sign of the interaction which indicates a promotive or repressive regulation $\mathrm{sgn}(k)\sim \mathrm{Bernoulli}(p_k)$ with $p_k\sim\mathrm{Beta}(0.5,0.5)$.
\end{itemize}
\begin{table}
\centering
\begin{tabular}{l|c|c|}
                                                                                                                      & 10X \textbf{Chromium} & \textbf{Drop-Seq}  \\ 
\hline
$p_\mathrm{outlier}$                                                                                  & 0.01                  & 0.01               \\
$\mu_\mathrm{outlier}$, $\sigma_\mathrm{outlier}$ & 3.0, 1.0              & 3.0, 1.0           \\
$\mu_\mathrm{lib}$, $\sigma_\mathrm{lib}$       & 6.0, 0.3              & 4.4, 0.8           \\
$\delta$, $\eta$                                                                    & 74, 8                 & 85, 8             
\end{tabular}
\caption{Technical noise parameters for 10X Chromium and Drop-Seq Single-Cell RNA sequencing platforms that is used for experiments in this work.}
\label{tab:sergio_noise_params}
\end{table}
For technical noise, we consider two different platforms: 10X Chromium and Drop-Seq. The noise parameters used are suggested in \citep{dibaeinia2020sergio}. These parameters are presented in \cref{tab:sergio_noise_params}.
\section{Training and OOD Distributions of Design Environments}\label{app:env_train_ood}
\subsection{Synthetic Design Environment}\label{app:synth_train_ood}
For training distribution, we make the following choices:
\begin{itemize}
    \item Prior over graphs $p(A)=p_\mathrm{ER}(k_{in}=3)$ is an Erdős–Rényi~\citep{erdHos1960evolution} with $3$ edges per node in expectation.
    \item Prior over parameters $p(\theta)= \gN(0, \sigma^2_\theta)$ where $\sigma^2_\theta$ is chosen such that marginal variance of each variable is 1~\citep{reisach2021beware}.
    \item Noise $\epsilon\sim \gN(0,\sigma^2_\epsilon)$ where $\sigma_\epsilon\sim \mathrm{InvGamma}(10,1)$.
\end{itemize}
For an OOD distribution, all the priors except for the parameter that undergoes distribution shift remain the same as during training. We define the following OOD environments and their corresponding distribution shifts:
\begin{enumerate}[label=\alph*]
     \item \textbf{Graphs:} $p(A)=p_\mathrm{SF}(k_{in}=3)$ is a Scale-Free~\citep{barabasi1999emergence} with $3$ edges per node in expectation.
    \item \textbf{Graphs+Mechanisms:} Prior over graphs is $p(A)=p_\mathrm{SF}(k_{in}=3)$ and prior over parameters $p(\theta)= \gN(0.1, \sigma^2_\theta)$ with $\sigma^2$ chosen as during training.
    \item \textbf{Graphs+Mech.+Noise:} Prior over graphs is $p(A)=p_\mathrm{SF}(k_{in}=3)$ and prior over parameters $p(\theta)= \gN(0.1, \sigma^2_\theta)$ with $\sigma^2$ chosen as during training. Noise $\epsilon\sim \mathrm{Gumbel}(0,\sigma_\epsilon)$ where $\sigma_\epsilon\sim \mathrm{InvGamma}(10,1)$.
    \item \textbf{Heteroskedastic Noise:} The causal model changes from \eqref{eq:anm} to \eqref{eq:anm_hk}.
    \item \textbf{Intervention Type:} The performed intervention in the environment changes from a $\mathrm{do}$ to a shift intervention.
    \item \textbf{Dimensionality $d$:} The dimensionality of the environment increases from training distribution ($d<10$).
\end{enumerate}

\subsection{Single-Cell Gene Regulatory Network Environment}\label{app:sergio_train_ood}
For training distribution, we make the following choices:
\begin{itemize}
    \item Prior over graphs $p(A)=p_\mathrm{ER}(k_{in}=3)$ is an Erdős–Rényi~\citep{erdHos1960evolution} with $3$ edges per node in expectation.
    \item Prior over mechanisms are as given in \cref{app:sergio_sim_params}.
    \item For technical noise, we consider the 10X Chromium platform whose parameters are given in \cref{tab:sergio_noise_params}.
\end{itemize}
For an OOD distribution, all the priors except for the parameter that undergoes distribution shift remain the same as during training. We define the following OOD environments and their corresponding distribution shifts:
\begin{enumerate}[label=\alph*]
     \item \textbf{Graphs:} $p(A)=p_\mathrm{SF}(k_{in}=3)$ is a Scale-Free~\citep{barabasi1999emergence} with $3$ edges per node in expectation.
    \item \textbf{(Technical) Noise:} The single-cell RNA sequencing platform changes from 10X Chromium to Drop-Seq, thereby changing the noise levels~\cref{tab:sergio_noise_params}.
    \item \textbf{Intervention Type:} The performed intervention in the environment changes from a gene knockout to a gene knockdown.
    \item \textbf{Noisy Interventions:} With a 10\% probability, the gene suggested by the policy for knockout is either does not happen, or there is an off-target that is knocked-out. We achieve this by flipping the one hot encoding of the intervention target labels with 10\% probability. But the history is only appended with the intervention sampled from the policy. Therefore, the policy has no knowledge of the noisy intervention.
\end{enumerate}
\section{Training Details}\label{app:training_details}
\subsection{Architecture Details}
We use the alternating attention based transformer for both the policy and the Q-function approximation. We maintain the same architecture for the transformer for both the policy and the Q-function, which we describe below.

For the transformer, we use a standard transformer block~\citep{vaswani2017attention} with 8 heads of self-attention. As our transformer has alternating attention, each layer has two such self-attention operations. Each self-attention is followed by a feedforward layer, whose dimension is set to $4*l$ where $l$ is the size of the state representation. We choose $l=32$. After $L$ layers of alternate attention, we perform max pooling over ordering of the data to obtain the state representation. The state representation is passed through a two hidden layer MLP with 128 hidden dimensions each and ReLU nonlinearity. 
\subsection{Hyperparameter Tuning}
REDQ~\citep{chen2021randomized} algorithm based on SAC~\citep{haarnoja2018soft} trains $M$ different Q-function networks and updates the gradients of each of them $G$ times before updating the policy. We treat both these quantities as hyperaprameters. All the parameters are updated with the Adam optimizer~\citep{kingma2014adam} and the learning rate is tuned. We list all the hyperparameters and the corresponding grid search in \cref{tab:hyperparams}.

\begin{table}[]
\caption{Hyperparameters used for training in CAASL.}
\label{tab:hyperparams}
\resizebox{\columnwidth}{!}{
\begin{tabular}{l|l|c|c|c}
                                                                                                                          &                                                                                         &                             & \textbf{\begin{tabular}[c]{@{}c@{}}Synthetic \\ Environment\end{tabular}}                 & \textbf{\begin{tabular}[c]{@{}c@{}}SERGIO\\ Environment\end{tabular}}                     \\
                                                                                                                          &                                                                                         & \textbf{Hyperameter Search} &                                                                                           &                                                                                           \\ \hline
\multirow{10}{*}{\textbf{\begin{tabular}[c]{@{}l@{}}Transformer \\ parameters\\ (History state\\  encoder)\end{tabular}}} & \begin{tabular}[c]{@{}l@{}}No. attention layers\\ (for policy, Q-Function)\end{tabular} &                             & 4                                                                                         & 3                                                                                         \\
                                                                                                                          & \begin{tabular}[c]{@{}l@{}}No. attention heads\\ (for policy, Q-Function)\end{tabular}  &                             & 8                                                                                         & 8                                                                                         \\
                                                                                                                          &                                                                                         & \multicolumn{1}{l|}{}       & \multicolumn{1}{l|}{}                                                                     & \multicolumn{1}{l}{}                                                                      \\
                                                                                                                          & $l$                                                                                     &                             & 32                                                                                        & 32                                                                                        \\
                                                                                                                          &                                                                                         & \multicolumn{1}{l|}{}       & \multicolumn{1}{l|}{}                                                                     & \multicolumn{1}{l}{}                                                                      \\
                                                                                                                          & Dropout (Policy)                                                                        &                             & 0.1                                                                                       & 0.1                                                                                       \\
                                                                                                                          &                                                                                         & \multicolumn{1}{l|}{}       & \multicolumn{1}{l|}{}                                                                     & \multicolumn{1}{l}{}                                                                      \\
                                                                                                                          & Pooling (Policy)                                                                        &                             & Max pool over samples                                                                     & Max pool over samples                                                                     \\
                                                                                                                          &                                                                                         & \multicolumn{1}{l|}{}       & \multicolumn{1}{l|}{}                                                                     & \multicolumn{1}{l}{}                                                                      \\
                                                                                                                          & Pooling (Q Function)                                                                    &                             & \begin{tabular}[c]{@{}c@{}}Max pool over samples, \\ sum pool over variables\end{tabular} & \begin{tabular}[c]{@{}c@{}}Max pool over samples, \\ sum pool over variables\end{tabular} \\ \hline
\multirow{4}{*}{\textbf{\begin{tabular}[c]{@{}l@{}}Decoder\\ parameters\end{tabular}}}                                    &                                                                                         & \multicolumn{1}{l|}{}       & \multicolumn{1}{l|}{}                                                                     & \multicolumn{1}{l}{}                                                                      \\
                                                                                                                          & \begin{tabular}[c]{@{}l@{}}Hidden sizes \\ (for policy and Q)\end{tabular}              &                             & (128, 128)                                                                                & (128, 128)                                                                                \\
                                                                                                                          &                                                                                         & \multicolumn{1}{l|}{}       & \multicolumn{1}{l|}{}                                                                     & \multicolumn{1}{l}{}                                                                      \\
                                                                                                                          & Non-linearity                                                                           &                             & ReLU                                                                                      & ReLU                                                                                      \\ \hline
\multirow{14}{*}{\textbf{\begin{tabular}[c]{@{}l@{}}REDQ/ SAC \\ training \\ parameters\end{tabular}}}                    &                                                                                         & \multicolumn{1}{l|}{}       & \multicolumn{1}{l|}{}                                                                     & \multicolumn{1}{l}{}                                                                      \\
                                                                                                                          & $M$                                                                                     & $\{2,3,5\}$                 & 5                                                                                         & 2                                                                                         \\
                                                                                                                          &                                                                                         & \multicolumn{1}{l|}{}       & \multicolumn{1}{l|}{}                                                                     & \multicolumn{1}{l}{}                                                                      \\
                                                                                                                          & $G$                                                                                     & $\{1, 3, 5\}$               & 1                                                                                         & 1                                                                                         \\
                                                                                                                          &                                                                                         & \multicolumn{1}{l|}{}       & \multicolumn{1}{l|}{}                                                                     & \multicolumn{1}{l}{}                                                                      \\
                                                                                                                          & $\gamma$                                                                                & $\{0.9, 0.95\}$             & 0.9                                                                                       & 0.95                                                                                      \\
                                                                                                                          &                                                                                         & \multicolumn{1}{l|}{}       & \multicolumn{1}{l|}{}                                                                     & \multicolumn{1}{l}{}                                                                      \\
                                                                                                                          & Buffer Size                                                                             & $\{10e6, 10e7\}$            & $10e7$                                                                                    & $10e6$                                                                                    \\
                                                                                                                          &                                                                                         & \multicolumn{1}{l|}{}       & \multicolumn{1}{l|}{}                                                                     & \multicolumn{1}{l}{}                                                                      \\
                                                                                                                          & Policy LR                                                                               & $\{0.01, 0.001\}$           & 0.001                                                                                     & 0.01                                                                                      \\
                                                                                                                          &                                                                                         & \multicolumn{1}{l|}{}       & \multicolumn{1}{l|}{}                                                                     & \multicolumn{1}{l}{}                                                                      \\
                                                                                                                          & Q-Function LR                                                                           & $\{3e-5, 3e-6\}$            & $3e-5$                                                                                    & $3e-6$                                                                                    \\
                                                                                                                          &                                                                                         & \multicolumn{1}{l|}{}       & \multicolumn{1}{l|}{}                                                                     & \multicolumn{1}{l}{}                                                                      \\
                                                                                                                          & $\tau$                                                                                  &                             & 0.01                                                                                      & 0.01                                                                                     
\end{tabular}
}
\end{table}
\section{Computational Resources}\label{app:comp_res}
We train all models on 3 40GB NVIDIA A100 GPU accelerators. We provide a wall time of 3 days, which results in a total computational budget of 216 GPU hours for each model. We also tune hyperparameters as outlined in \cref{tab:hyperparams} for both environments, resulting in a total usage of 70,000 hours. For testing, we just rollout the policy on a CPU, which can be completed in seconds.

\section{Licenses}\label{app:licenses}
For the single-cell gene simulator, we make use of the publicly available repository which is released under \href{https://github.com/PayamDiba/SERGIO?tab=GPL-3.0-1-ov-file#readme}{GPL-3.0 License}
For the reward model AVICI, we make use of the publicly released code and trained models. These are released under \href{https://github.com/larslorch/avici/tree/main?tab=MIT-1-ov-file#readme}{MIT License}. 
For baselines, we use the DiffCBED open source repository, which is released under \href{https://github.com/yannadani/DiffCBED?tab=MIT-1-ov-file#readme}{MIT License}.
\section{Full Results}
\subsection{Results on Zero-Shot OOD Generalization to Higher Dimensions}
The results for Zero-shot OOD generalization to problems of higher dimensions is available in \cref{fig:linanm_multiple_d,fig:sergio_d}.
\begin{figure}
    \centering
    \includegraphics[width=\linewidth]{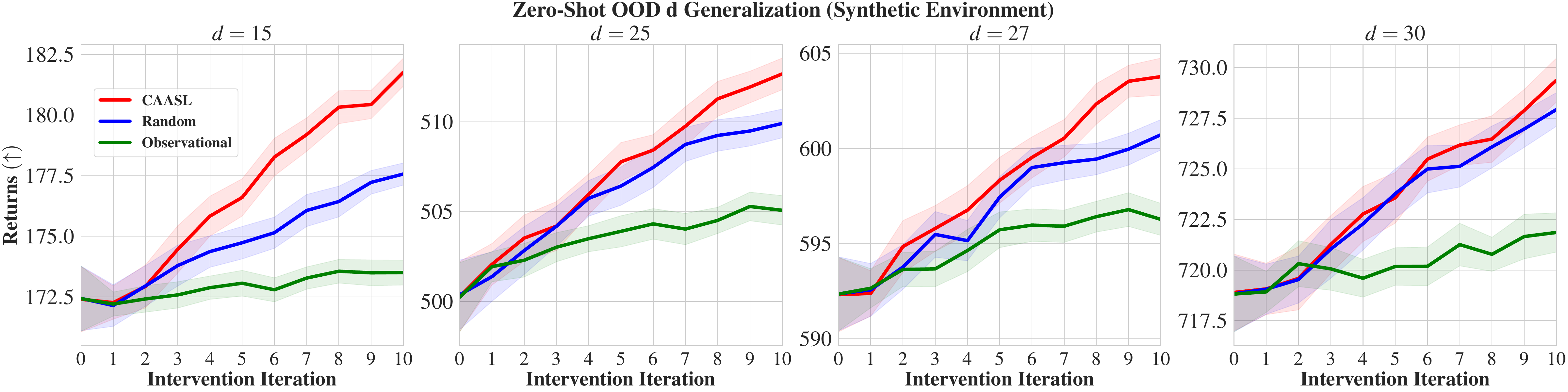}
    \caption{Results of zero-shot OOD generalization when dimensionality of the data increases in the synthetic environment.  Results are performed on 100 random test environments. Shaded area represents 95\% CI. }
    \label{fig:linanm_multiple_d}
\end{figure}
\begin{figure}
    \centering
    \includegraphics[width=0.5\linewidth]{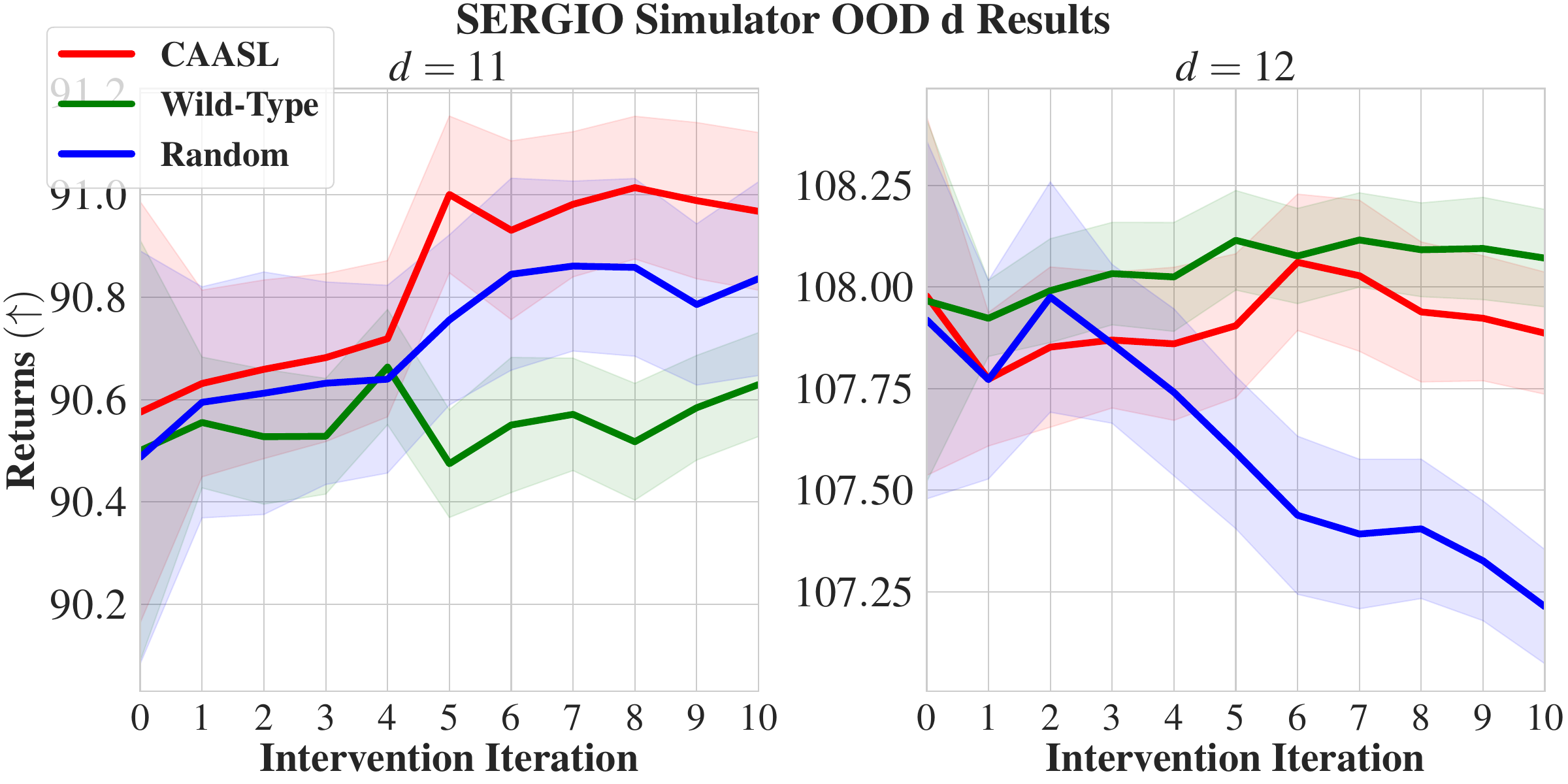}
    \caption{Results of zero-shot OOD generalization when dimensionality of the data increases in the SERGIO environment. We notice that the random baseline is very competitive. We hypothesize that the symmetries encoded in the policy, which are crucial for generalization, might not be so relevant in this setting due to high amount of missing data. Results are performed on 100 random test environments. Shaded area represents 95\% CI. }
    \label{fig:sergio_d}
\end{figure}
\subsection{Results on all Metrics}\label{app:all_res}
Herein we include all the results that correspond to other metrics omitted in the main text. In particular, apart from returns, we measure Structural Hamming Distance (SHD), Area under Precision Recall Curve (AUPRC), and Edge F1 (Edge F1) score. These additional metrics are defined as follows:
\begin{itemize}
    \item \textbf{SHD}: Structural Hamming Distance (SHD) measures the hamming distance between graphs. In particular, it is a measure of number of edges that are to be added, removed or reversed to get the ground truth from the estimated graph. Since we have a posterior distribution $q(\hat{A}\mid h_t)$ over graphs, we measure the \emph{expected} SHD: 
    \begin{equation*}
    \text{SHD} \coloneqq \E_{\hat{A}\sim q(\hat{A}\mid h_t)}[\mathrm{SHD}(\hat{A}, A^{{GT}})] \approx \frac{1}{100}\sum_{i=1}^{100}[\mathrm{SHD}(\hat{A}^{(i)}, A^{{GT}})]~~~~,\text{with}~~~\hat{A}^{(i)}\sim q(\hat{A}\mid h_t)
\end{equation*} where $A^{GT}$ is the ground-truth causal graph. 
\item \textbf{Edge F1}: It is F1 score of each edge being present or absent in comparison to the true edge set, averaged over all edges.
\item \textbf{AUPRC}: It is the area under the precision recall curve obtained by thresholding the edge probabilities of the amortized graph posterior $q(\hat{A}\mid h_t)$.
\end{itemize}
\begin{figure}
    \centering
    \includegraphics[width=\linewidth]{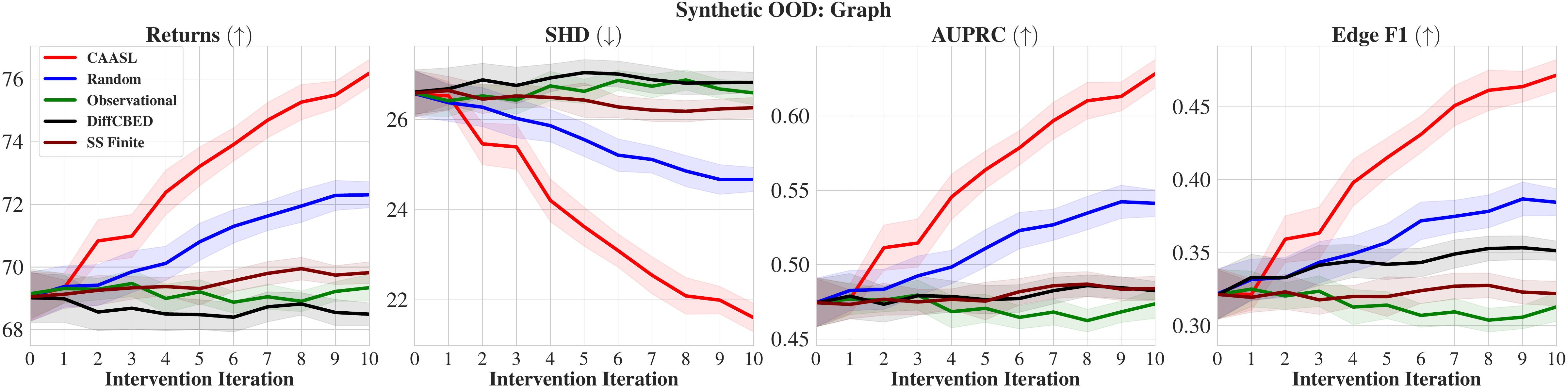}
    \caption{Results of zero-shot OOD graph setting with various intervention strategies on 100 random test environments. Shaded area represents 95\% CI. }
\end{figure}
\begin{figure}
    \centering
    \includegraphics[width=\linewidth]{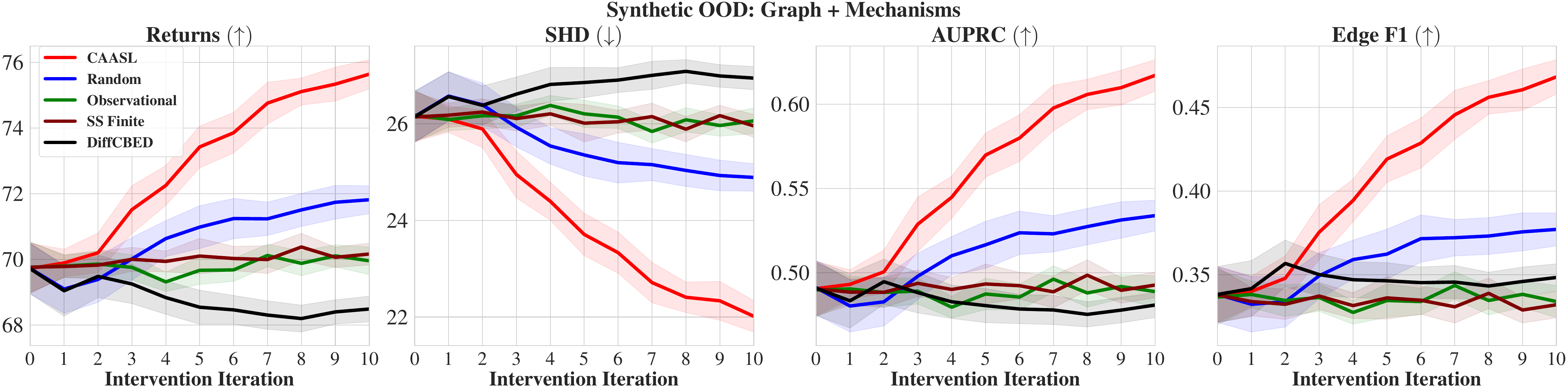}
    \caption{Results of zero-shot OOD graph and mechanisms setting with various intervention strategies on 100 random synthetic test environments. Shaded area represents 95\% CI. }
\end{figure}
\begin{figure}
    \centering
    \includegraphics[width=\linewidth]{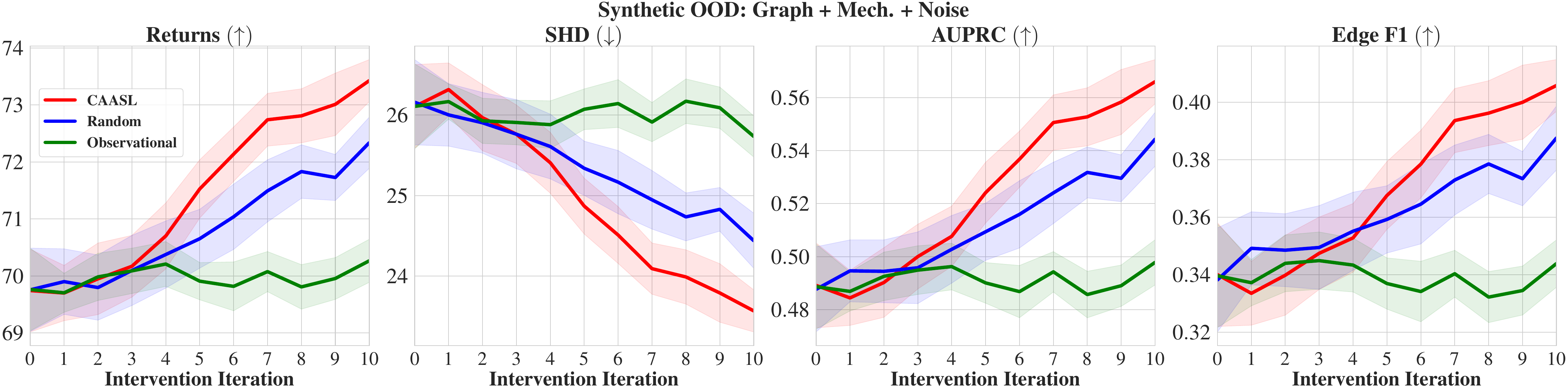}
    \caption{Results of zero-shot OOD graph, mechanisms and noise setting with various intervention strategies on 100 random synthetic test environments. Shaded area represents 95\% CI. }
\end{figure}
\begin{figure}
    \centering
    \includegraphics[width=\linewidth]{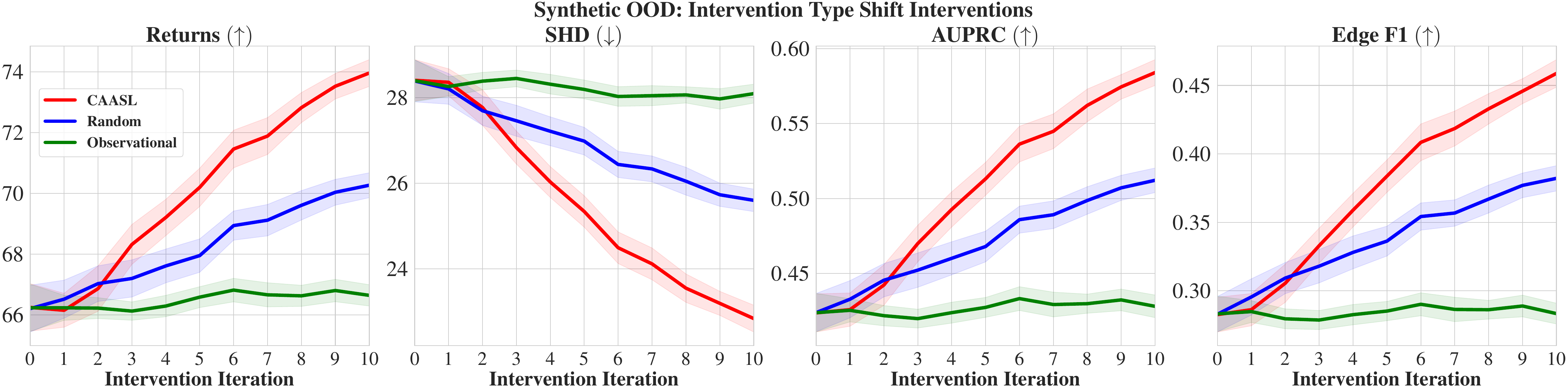}
    \caption{Results of zero-shot OOD intervention type setting with various intervention strategies on 100 random synthetic test environments. Shaded area represents 95\% CI. }
\end{figure}
\begin{figure}
    \centering
    \includegraphics[width=\linewidth]{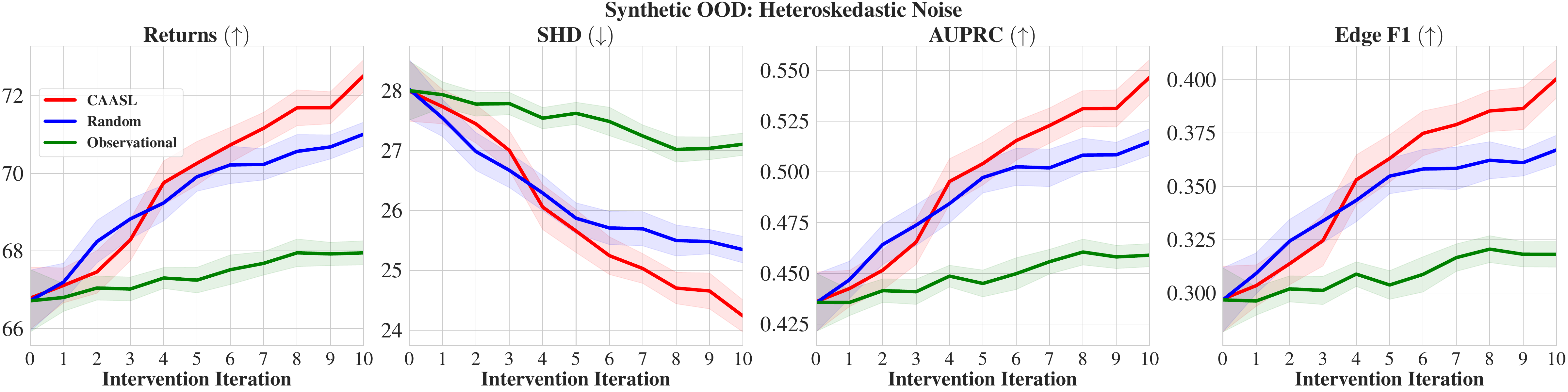}
    \caption{Results of zero-shot OOD heteroskedastic noise setting with various intervention strategies on 100 random synthetic test environments. Shaded area represents 95\% CI. }
\end{figure}
\begin{figure}
    \centering
    \includegraphics[width=\linewidth]{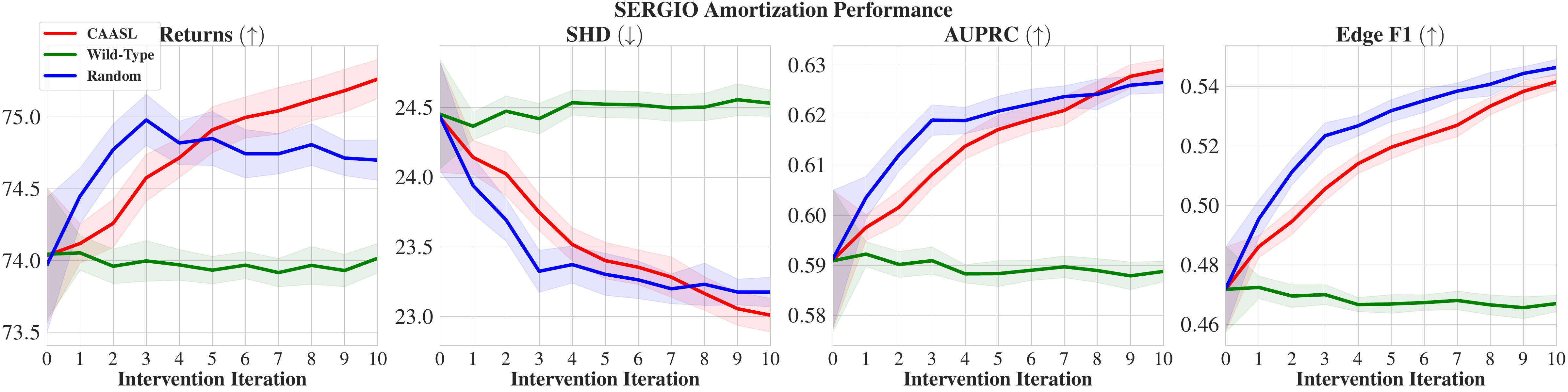}
    \caption{Results of  amortization with various intervention strategies on 100 random SERGIO test environments. Shaded area represents 95\% CI. }
\end{figure}
\begin{figure}
    \centering
    \includegraphics[width=\linewidth]{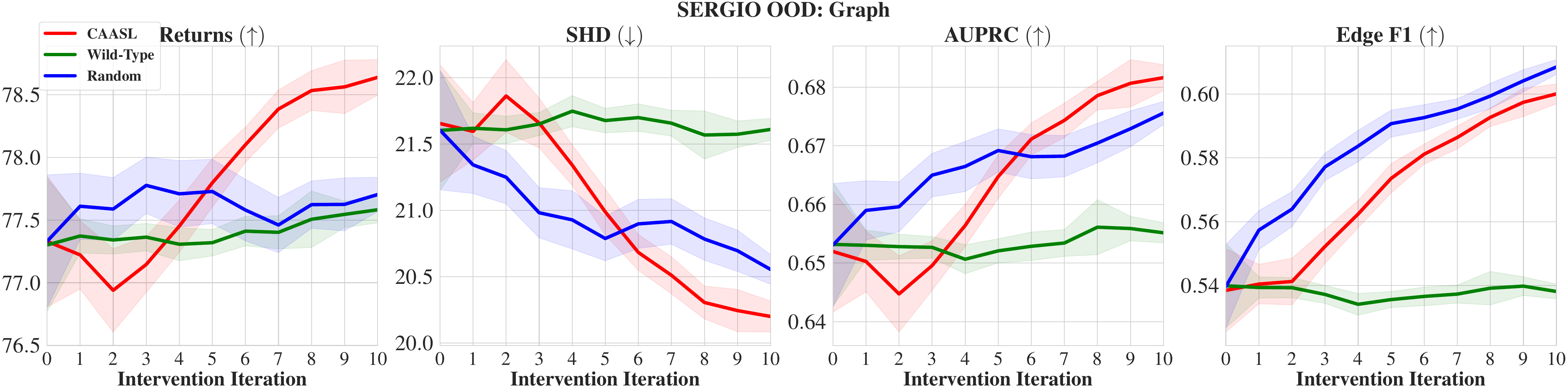}
    \caption{Results of zero-shot OOD graph setting with various intervention strategies on 100 random SERGIO test environments. Shaded area represents 95\% CI.}
\end{figure}
\begin{figure}
    \centering
    \includegraphics[width=\linewidth]{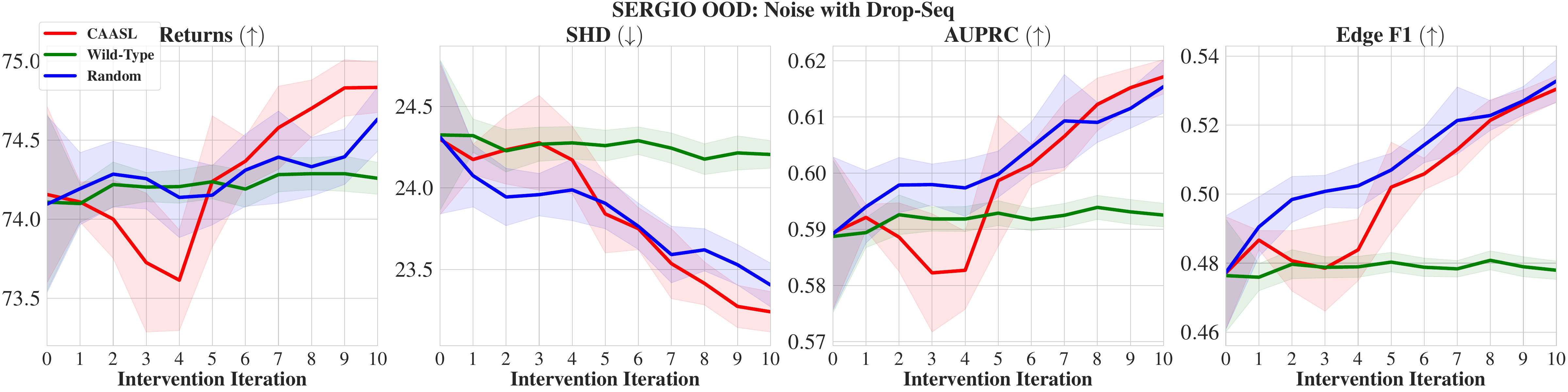}
    \caption{Results of zero-shot OOD scRNA-seq platform and their noise setting with various intervention strategies on 100 random SERGIO test environments. Shaded area represents 95\% CI. }
\end{figure}
\begin{figure}
    \centering
    \includegraphics[width=\linewidth]{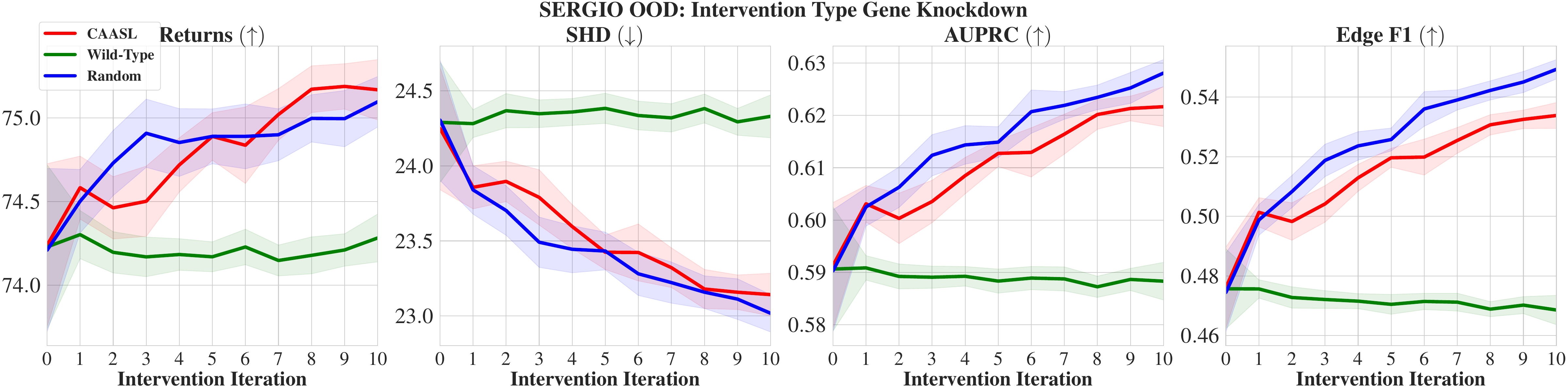}
    \caption{Results of zero-shot OOD intervention type changing to gene knockdown with various intervention strategies on 100 random SERGIO test environments. Shaded area represents 95\% CI. }
\end{figure}
\begin{figure}[!]
    \centering
    \includegraphics[width=\linewidth]{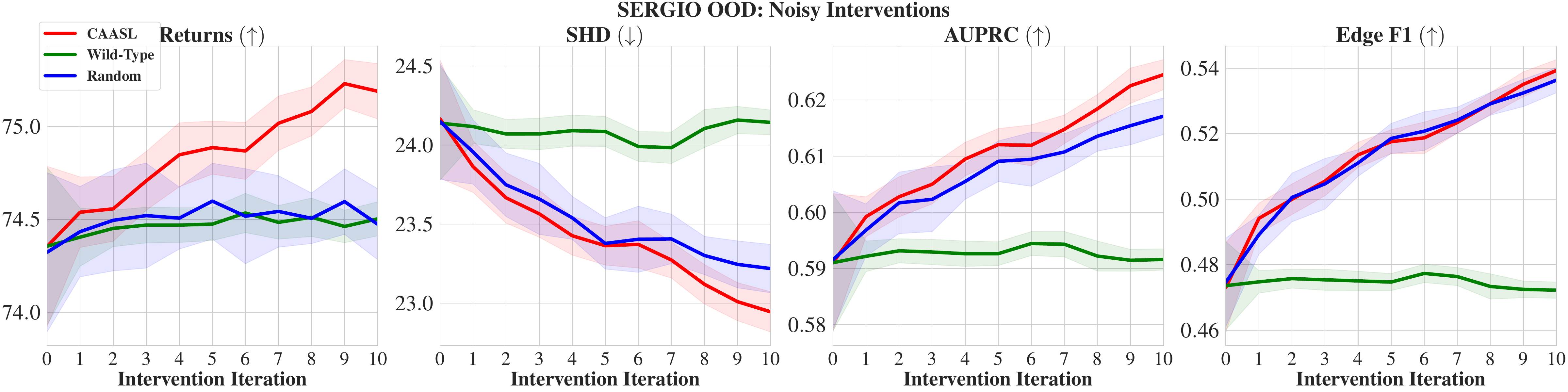}
    \caption{Results of zero-shot OOD noisy gene knockouts with various intervention strategies on 100 random SERGIO test environments. Shaded area represents 95\% CI. }
\end{figure}

\end{document}